\pdfoutput=1
\documentclass[11pt]{article}

\usepackage{ACL2023}

\usepackage{times}
\usepackage{latexsym}
\usepackage{graphicx}
\usepackage{algorithm2e}
\RestyleAlgo{ruled}
\usepackage{booktabs}
\usepackage{enumitem}
\usepackage{amsmath, amsfonts}
\usepackage{bm}
\usepackage{arydshln}
\usepackage{subfigure}
\usepackage{multirow}
\usepackage{tcolorbox}
\usepackage{svg}
\usepackage{amssymb}% http://ctan.org/pkg/amssymb
\usepackage{pifont}% http://ctan.org/pkg/pifont
\newcommand{\cmark}{\ding{51}}%
\newcommand{\xmark}{\ding{55}}%
\newcommand{\bluetext}{\textcolor{blue}}%
\newcommand{\redtext}{\textcolor{red}}
\newcommand{\dlb}{\textit{domain-label bias}}

\usepackage[T1]{fontenc}
\usepackage[utf8]{inputenc}

\usepackage{microtype}

\usepackage{inconsolata}

\usepackage{todonotes}
\newcommand\eg{\textit{e.g.}}
\newcommand\ie{\textit{i.e.}}

\title{%Random Text Calibration: \\on 
Mitigating Label Biases for In-context Learning
}

\author{Yu Fei$^{\dagger1}$, Yifan Hou$^{*2}$, Zeming Chen$^{*3}$, Antoine Bosselut$^3$\\
$^1$UC Irvine, $^2$ETH Zurich, $^3$NLP Lab, IC, EPFL, Switzerland\\ 
\texttt{yu.fei@uci.edu,} \texttt{yifan.hou@inf.ethz.ch,}\\
\texttt{\{zeming.chen, antoine.bosselut\}@epfl.ch}}

\begin{document}
\maketitle
\def\thefootnote{*}\footnotetext{Equal contribution.}\def\thefootnote{\arabic{footnote}}
\def\thefootnote{$\dagger$}\footnotetext{Work done while interning at EPFL.}\def\thefootnote{\arabic{footnote}}
\begin{abstract}
% Ver 1: from the angle of domain-label bias
% Recent work demonstrates that in-context learning performance is sensitive to a variety of design settings, such as the choice and order of the in-context examples. In this work, we show that the word distribution of the input text domain can also bias the model toward predicting certain label names. On many tasks, such \emph{domain-label} bias restricts GPT-3 to a chance-level performance regardless of the choice of in-context examples. To mitigate domain-label bias and also other biases introduced by the design setting, we first estimate the model's prior on predicting each label name using random words sampled from the domain. Then when making predictions, we calibrate the model's output probabilities using the prior. Our \emph{random text calibration} improves the in-context learning performance of GPT-J and GPT-3 on a wide range of classification tasks. We achieve an average improvement of up to 37\% (GPT-J) and 32\% (GPT-3) relative in Macro-F1 on tasks with large domain-label biases. Furthermore, our results generalize to settings where the model is instruction-tuned and/or provided with manually designed task instructions. From a broader view, the existence of domain-label bias motivates a more careful dataset-collecting process and designing of experiments to control the confounders for understanding the general behaviors of in-context learning.

% Antoine re-write v2:
Various design settings for in-context learning (ICL), such as the choice and order of the in-context examples, can bias a model toward a particular prediction without being reflective of an understanding of the task. While many studies discuss these design choices, there have been few systematic investigations into categorizing them and mitigating their impact. In this work, we define a typology for three types of label biases in ICL for text classification: \textit{vanilla-label bias}, \textit{context-label bias}, and \textit{domain-label bias} (which we conceptualize and detect for the first time). 

Our analysis demonstrates that prior label bias calibration methods fall short of addressing all three types of biases. Specifically, domain-label bias restricts LLMs to random-level performance on many tasks regardless of the choice of in-context examples.  To mitigate the effect of these biases, we propose a simple bias calibration method that estimates a language model's \textit{label bias} using random in-domain words from the task corpus. After controlling for this estimated bias when making predictions, our novel \textit{domain-context calibration} significantly improves the ICL performance of GPT-J and GPT-3 on a wide range of tasks. The gain is substantial on tasks with large domain-label bias (up to 37\% in Macro-F1). Furthermore, our results generalize to models with different scales, pretraining methods, and manually-designed task instructions, showing the prevalence of label biases in ICL. Our codes are available at \texttt{\url{https://github.com/fywalter/label-bias}}.
\end{abstract}

%%%%%%%%%%%%%%%%%%%%%%%%%%%%%%%%%%%%%%%%%%%%%%%%%%%%%%%%%%%%%%%%%%%
%%%%%%%%%%%%%%%%%%%%%%%%%%%%%%%%%%%%%%%%%%%%%%%%%%%%%%%%%%%%%%%%%%%
%%%%%%%%%%%%%%%%%%%%%%%%%%%%%%%%%%%%%%%%%%%%%%%%%%%%%%%%%%%%%%%%%%%
\section{Introduction}
% Large LLMs can perform unseen tasks simply by conditioning on a context prompt that consists of a few training example-label pairs \citep{brown2020language}. However, such in-context learning ability is highly sensitive to specific settings, such as the ordering of the in-context examples and the label word frequencies in the pre-training corpus \citep{liu2021makes, lu2021fantastically, zhao2021calibrate}. 

Large language models (LLMs) can perform unseen tasks by conditioning on a context prompt that consists of a few training example-label pairs \citep{brown2020language}. However, such in-context learning ability is highly sensitive to various design settings, such as the choice \citep{liu2021makes} and order \citep{lu2021fantastically} of the in-context samples. Recently, \citet{zhao2021calibrate} showed that the instability of ICL largely arises from the fact that these design settings bias the model toward predicting certain answers (\eg, LLMs often predict the label of the last in-context example). As a result, the sensitivity of the results in ICL studies calls for a systematic discussion of biases in ICL and new methods to properly categorize, detect, and comprehensively mitigate various types of biases.
% introduced by different sources. % various biases introduced by different contexts.%, and thus improve the performance and robustness of ICL.

%there hasn't been a systematic discussion of the biases in ICL to 1) categorize existing findings, 2) detect new biases, and 3) develop methods to comprehensively handle different types of biases to improve the performance and robustness of ICL.

\begin{figure}[t]
\centering
\includegraphics[width=\columnwidth]{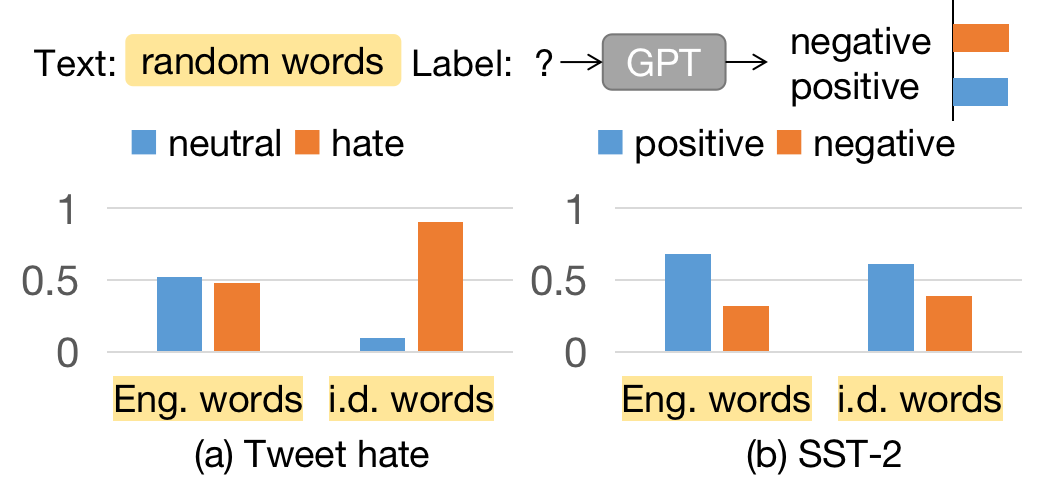}
% \caption{On two datasets, we compute the GPT-J's prior on each label using random English words (Eng. words) and random in-domain words sampled from the target dataset (i.d. words). Different from on SST-2, on TweetEval-hate, seeing random in-domain words severely biases the model to predict label hate, while seeing random English words shows no such preference.} 
\caption{Illustration of domain-label bias. (a) On a Twitter hate speech detection dataset (TweetEval-hate; \citealp{barbieri-etal-2020-tweeteval}), the model is severely biased toward predicting label \textit{hate} when random in-domain (i.d.) words from the dataset are provided as input. Random English (Eng.) words show no such preference. (b) On a movie review dataset, SST-2 \citep{socher2013recursive}, no such a bias is observed.} 
\label{fig:domain_label_bias}
\end{figure}

% In this work, we conduct a thorough investigation of the bias problem in ICL in the context of text classification. We start by defining a typology consisting of three types of \textit{label biases} (the model's undesirable preference on certain labels): \textit{vanilla-label bias}, \textit{context-label bias}, and \textit{domain-label bias}. \textit{Vanilla-label bias} captures the model's uncontextual preference on the label names, which can be caused by different frequencies of label names in the pre-training corpus \citep{zhao2021calibrate}. \textit{Context-label bias} summarizes the effects of the context prompt. For example, it considers the bias caused by the order of the in-context examples. Finally, \textit{domain-label bias} captures the effects of the task corpus on the model's prediction, which is missing in existing discussions of ICL.

In this work, we conduct a thorough investigation of biases in ICL for text classification. We start by defining a typology of three types of \textit{label biases} (the model's undesirable preference toward certain label names): \textit{vanilla label bias}, \textit{context-label bias}, and \textit{domain-label bias}. What we term \textit{vanilla label bias} captures the model's non-contextualized preference for the label names (\eg, the common token bias mentioned by \citet{zhao2021calibrate} caused by different frequencies of label names in the pretraining corpus). \textit{Context-label bias} summarizes the effects of the context prompt (\eg, LLMs tend to prefer the majority and last label of the in-context examples). Finally, \textit{domain-label bias} captures the effects of the task corpus on the model's predictions.

\begin{figure}[t]
\centering
\includegraphics[width=\columnwidth]{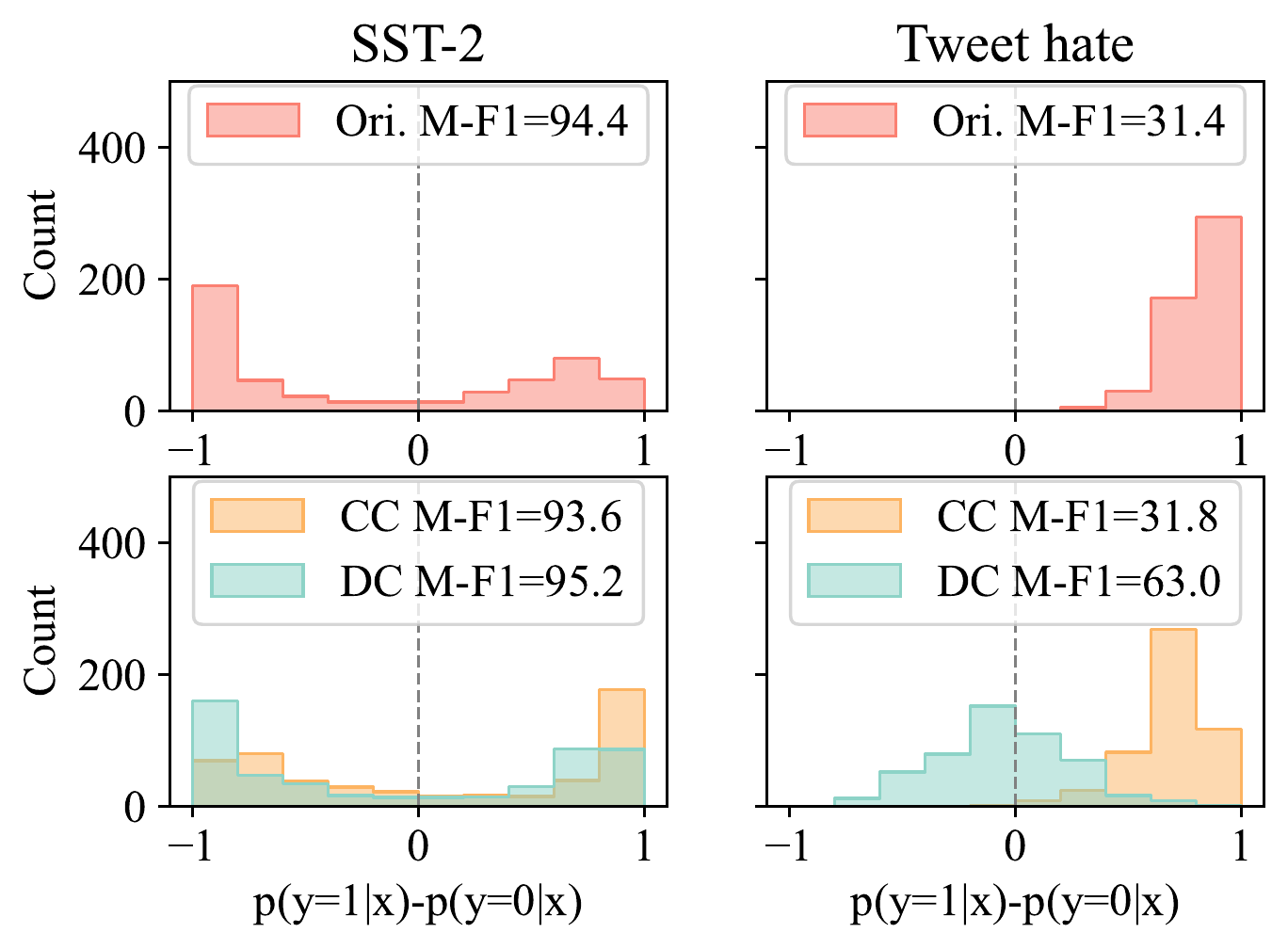}
\caption{Prediction distributions of GPT-J. On SST-2, GPT-J has a balanced prediction distribution. However, on TweetEval-hate \citep{barbieri-etal-2020-tweeteval}, the original (Ori.) and calibrated model (with contextual calibration, CC) predict most inputs as hateful. Our domain-context calibration (DC) largely mitigates such bias and substantially boosts the performance (Macro-F1).}
\label{fig:prediction_dist}
\end{figure}

We show that domain-label biases significantly affect a model's prediction in ICL.  For example, on a hate detection task with two nearly balanced classes, we observe that when random words are sampled from the dataset and provided as input, the model is severely biased towards predicting the label \textit{hate} (Fig.~\ref{fig:domain_label_bias}(a)). When random English words are provided as part of the input, no such effect is observed. More importantly, on many tasks with large domain-label bias, LLMs achieve no better than random performance, regardless of the choice of in-context examples (Fig.~\ref{fig:prediction_dist}). Moreover, we find that existing bias mitigation methods, such as Contextual Calibration (CC; \citealp{zhao2021calibrate}) do not combat this effect. %\antoine{I don't like the flow of the previous paragraph. I'll come back}

To this end, we propose Domain-context Calibration (DC) to mitigate {label biases} in ICL. DC first estimates the effects of different {label biases} holistically using random words sampled from the task corpus. Specifically, we compute the probabilities assigned by the model to each label using random in-domain words as the task input (with optional real in-context learning examples prepended). Using random words limits the semantic meaning of the input, allowing us to estimate the {vanilla-label} and {context-label biases} while using in-domain words accounts for the effect of the task corpus. Then, at inference time, we use this label bias estimate to calibrate the model's output probabilities.

We evaluate the impact of DC on 24 classification datasets, showing that DC improves the average ICL performance of GPT-J \cite{gpt-j} and GPT-3 by 20\% and 18\%. We observe substantial gains on tasks with large domain-label bias (up to 37\% in Macro-F1). DC also benefits models with different scales, instruction-tuning (\eg, Instruct-GPT, \citealp{ouyang2022training}), and provided with task instructions. 
% With a systematic ablation study, we identify two drawbacks of CC: 1) the pre-defined content-free tokens like ``N/A'' may lead to biases; 2) instead of using only a single content-free token, using content-free texts whose length matches the average input length of the dataset is beneficial. 
Finally, we show that DC improves the zero-shot prompting performance of smaller models like RoBERTa \citep{liu2019roberta}, demonstrating that label bias can be mitigated in prompt-based frameworks beyond ICL.

Overall, our work proposes a new typology of \textit{label biases} in prompt-based methods and a simple method for mitigating them. When studying ICL on a diverse collection of datasets, the results on datasets with severe \textit{label bias} can obfuscate the actual behaviors of the model. Thus, rigorous design for dataset selection (that accounts for confounders) and  fine-grained analysis of individual datasets are essential for effectively studying ICL.
% provides a more solid starting point for understanding the in-context learning ability of the LLMs.
% From a broader view, the existence of domain-label bias suggests that people need to be more careful in selecting datasets for understanding in-context learning. When the PLM always predict one label no matter what input is given, in-context learning is not sensitive to gold labels seems not to be the most reasonable conclusion to draw.

% Overall, our contributions are three-folded: 1) We show that the word distributions of the datasets can lead to severe biases in in-context learning; 2) We propose a simple calibration method that largely mitigates such domain-label bias; 3) We identified two drawbacks of the existing contextual calibration method that are fixed automatically with our random text calibration.
\begin{figure}[t]
\centering
\includegraphics[width=\columnwidth]{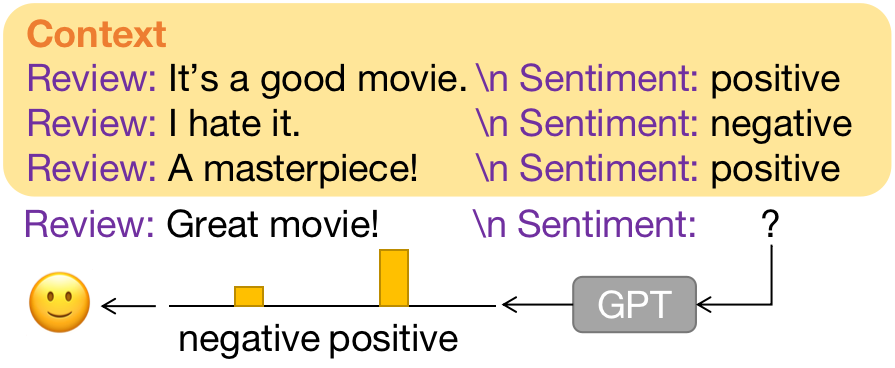}
\caption{Illustration of few-shot in-context learning for sentiment classification. The \textcolor{orange}{context prompt} consists of \textcolor{violet}{task-specific templates} and a few example-label pairs.}
\label{fig:in-context}
\end{figure}

\section{Categorizing Label Biases in ICL}
In this paper, we focus on in-context learning (ICL; Fig.~\ref{fig:in-context}) for classification tasks. Formally, we consider a dataset of examples $\{x_i, y_i\}$ where $x_i$ are text inputs and each $y_i$ can be mapped to a verbalization in a label name set $\mathcal{L}$. We assume each class has one label name. For example, in a sentiment task, $\mathcal{L}$ could be composed of \textit{positive} and \textit{negative} as label names. Given a context prompt $C$ consisting of a few labeled examples and an input text $x_i$, the model $M$ determines the label of $x_i$ by computing: $\mathop{\arg\max}_{y\in\mathcal{L}}P_M(y|x_i,C)$. Using this notation, we define our typology of label biases based on the mathematical formulation of ICL.
\subsection{A Typology of Label Biases}
\label{ssec:typology}
\begin{figure}[t]
\centering
\includegraphics[width=\columnwidth]{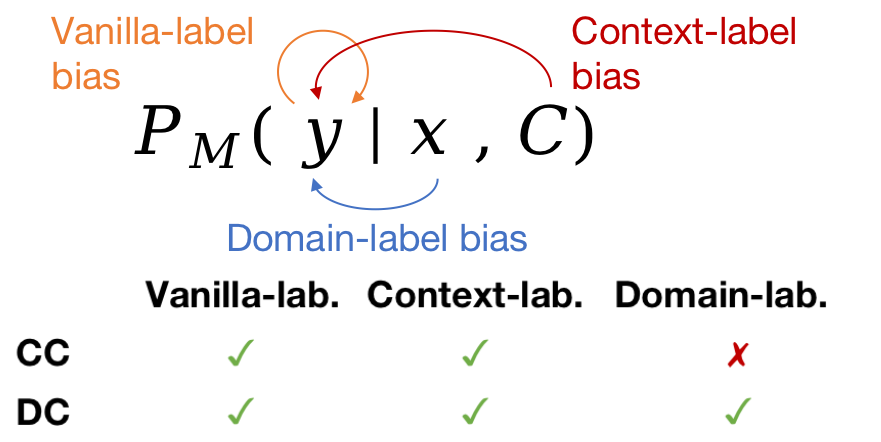}
\caption{Three types of biases in ICL. Contextual calibration (CC) considers only the first two types, while our domain-context calibration (DC) handles all of them.}
\label{fig:bias_source}
\end{figure}

To perform a classification task, a model needs to learn the underlying text-label mapping, i.e., $P(y|x)$. In supervised learning, such mapping is learned by optimizing the model using the training data. In ICL, on the other hand, the model is fixed, and it determines the label of a text by computing the probabilities of predicting the label names $P_M(y|x,C)$. Notice that there are three components involved in the inference: the label name $y$, the text $x$ from a specific task corpus, and the context $C$. Accordingly, as shown in Fig.~\ref{fig:bias_source}, we can define three types of label biases that lead to a discrepancy between $P_M(y|x,C)$ and $P(y|x)$.

% To achieve good performance for a classification tasks, a model needs to learn the underlying text-label mapping, i.e. $P(y|x)$. In supervised learning, such mapping is learned by optimizing the model using the training data. In ICL, on the other hand, the model is fixed, and it determine the label of a text by computing the probabilities of predicting the label names. This process relies on the model's prior knowledge of the task, i.e. the text-label association learned from pre-training, and conditioning on a context $C$ consisting of $(x,y)$ pairs that provide information of the desired text-label mapping. Both of these two aspects can result in \textit{label biases}, which we categorize into three types (Fig.~\ref{fig:bias_source}).

\paragraph{Vanilla-label bias} pertains to the uncontextual preference of the model towards predicting certain label names. One possible cause is the pre-training term frequencies of the label names. \citet{zhao2021calibrate} reported a high correlation between the frequency of the DBPedia dataset label names and the rate at which GPT-3 predicts those labels.

\paragraph{Context-label bias} summarizes the effect of the context prompt. %Unlike supervised learning, conditioning on text-label pairs does not provide any guarantee of the learned text-label mapping. 
With in-context learning, the model ``learns'' from a few examples, and the learning is particularly sensitive to seemingly arbitrary decisions such as the order of the in-context examples \citep{lu2021fantastically} and the task template used to map the example to text that the model can process \citep{mishra2021reframing, holtzman2021surface}.

%noisy to and may focus on simple heuristic associations in the context examples, rather than learning a robust mapping. Also, many aspects of the conditioning process introduce extra noise to the learning process. The perturbations caused by permuting the order of the in-context examples \citep{lu2021fantastically} and different task templates \citep{mishra2021reframing, holtzman2021surface} are two examples of such noise.

\paragraph{Domain-label bias} captures the effect of the task corpus. Beyond the text-label association demonstrated in the in-context examples, the model also relies on its prior knowledge of the task when making predictions. We show that the association of words to the label names learned from pre-training is a potential pitfall and discuss domain-label bias in more detail in the next section.

\section{Domain Label Bias}\label{sec:domain-label_bias}
% To illustrate how the domain of a task can induce label bias, consider a case where a model must detect the sentiment of a text that is sarcastic. An example text such as ``Nice perfume! How long did you marinate in it?'' contains positive words, but the sentiment is negative based on the context. However, as sarcastic texts may be sparse in the natural corpus, a language model would associate most texts containing positive words with positive sentiment while the actual sentiment is determined based on the context. \TODO{medical analysis example}
{To illustrate how the domain of a task can induce label bias, consider a case where an LLM predicts whether a patient is \textit{sick} or \textit{healthy} based on some medical descriptions. Because medical descriptions are associated more often with people having health problems in the natural corpus, frequently used words in such documents are likely to have a stronger correlation with \textit{sick} than \textit{healthy}, leading to a systematic bias in the model's predictions.}

Supporting this intuition, we find that for many datasets, conditioning on random words from the dataset examples biases the model toward predicting certain label names. For example, in the hate speech detection task depicted in Fig.~\ref{fig:domain_label_bias}, we compute the model's preference (prior) on both label names given random words as the input. A model such as GPT-J has no preference for either of the classes (\textit{neutral} v.s. \textit{hate}) given random English words, but given random in-domain words sampled from the dataset, the label priors shift dramatically, becoming 0.95 (\textit{hate}) v.s. 0.05 (\textit{neutral}). %On a sentiment dataset, we do not observe such a difference. 
% We call such bias domain-label bias as it is the distribution of words from the target tasks that makes the model prefer certain label names.

\begin{figure*}[t]
\centering
\includegraphics[width=0.49\textwidth]{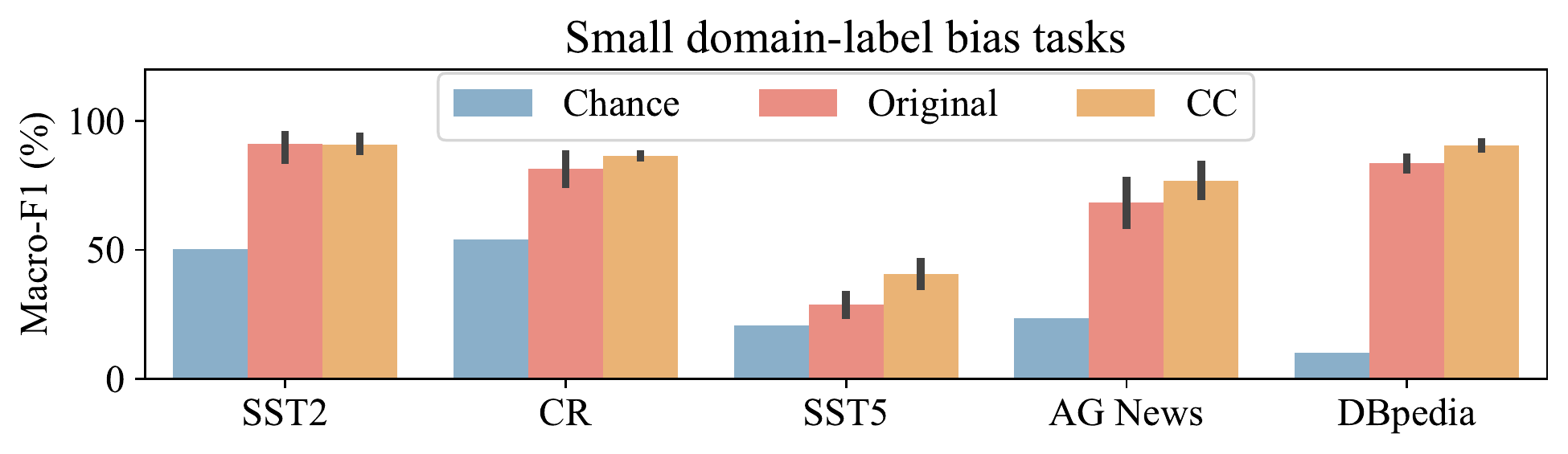}
\includegraphics[width=0.49\textwidth]{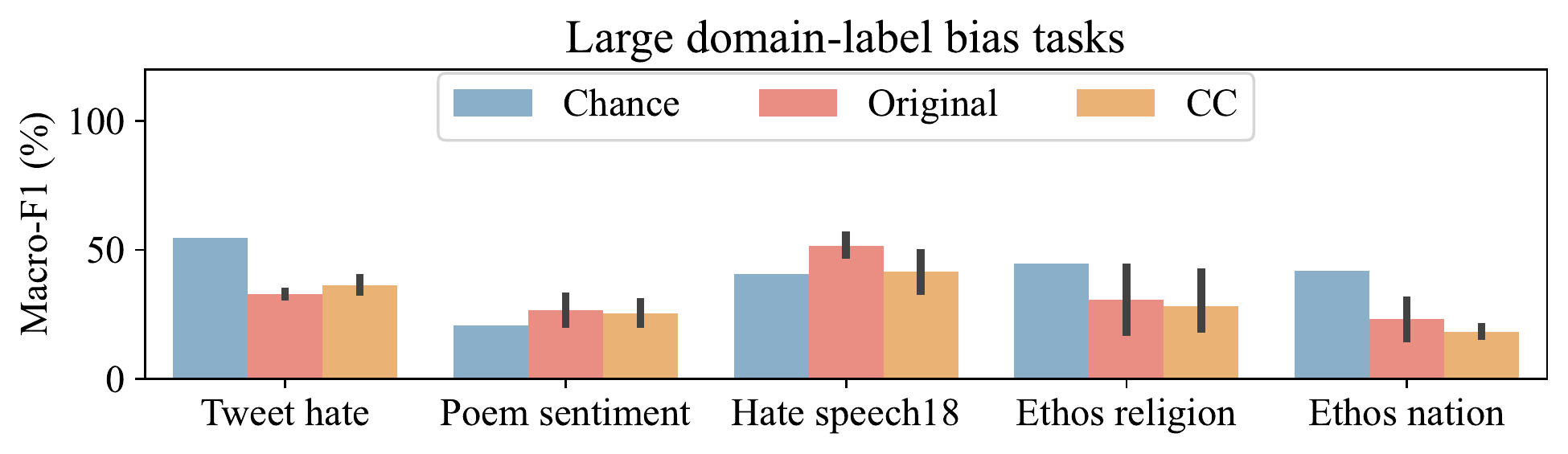}
% \caption{Top: the \textit{domain-label bias} of GPT-J on all the 24 evaluation datasets. Bottom: Despite having strong performance on datasets with small \dlb, GPT-J (8-shot) shows trivial improvements over the random baseline on datasets with large \dlb, where applying CC can lead to even worse performance.}
% \caption{Despite having strong performance on datasets with small \dlb, GPT-J (8-shot) shows trivial improvements over the random baseline on datasets with large \dlb, where applying CC can lead to even worse performance.}
\caption{In-context learning performance of GPT-J on datasets with varying levels of domain-label bias. On datasets with large domain-label bias, GPT-J under-performs the random baseline. More importantly, on these datasets, applying contextual calibration (CC) is not helpful and can often lead to lower performance.}
\label{fig:many_datasets}
\end{figure*}

Motivated by this experiment, we quantify the domain-label bias of a model for a particular task using the distance of the model's priors estimated using random English words and in-domain words. To make the measure more comparable on tasks with different numbers of classes, we define the following metric:%metmeasureric: %use the $L_1$ distance as the metric so that the measure is always in $[0,1]$:
\begin{equation}\label{eq:domain_label}
    bias=\frac{1}{2}\sum_{y\in \mathcal{L}} \Big|P_M(y|x_{Eng.})-P_M(y|x_{i.d.})\Big|,
\end{equation}
where $x_{Eng.}$ and $x_{i.d.}$ correspond to $L$ random English or random in-domain words and $L$ is the average text length of the dataset. 

We find that datasets\footnote{We note that domain-label bias is model-dependent. However, we observe a high correlation of domain-label bias between LLMs in general (see App.~\ref{sec:correlation}). Also, by definition, domain-label bias depends on the task formulation, particularly the choice of label names, which we discuss in \S~\ref{sec:discussion}.} exhibit different levels of domain-label bias (see Fig.~\ref{fig:domain-label_dataset} in App.~\ref{sec:domain-label_bias_all}). More importantly, LLMs behave distinctively on datasets with small and large domain-label bias. As shown in Fig.~\ref{fig:many_datasets}, while GPT-J performs competitively on datasets with small domain-label bias, it rarely outperforms the random baselines on large-bias datasets, indicating that domain-label bias significantly affects ICL. Contextual calibration, which only considers vanilla-label bias and context-label bias, fails to handle domain-label bias.
\section{Domain-context Calibration}\label{sec:rc}
%We have seen that ICL is affected by three potential types of label biases. 
In this section, we propose Domain-context Calibration (DC), which mitigates the effects of the multiple label biases of our typology (\S~\ref{ssec:typology}). Following contextual calibration (CC; \citealp{zhao2021calibrate}), we estimate the overall label bias of a model with respect to a task by estimating the label probabilities on a \textit{content-free} example text. However, unlike CC, which uses a single, seemingly content-free token (\eg, ``N/A'') to approximate the label bias, we use random words sampled from the unlabeled evaluation dataset as the content-free text. Then, for all examples we classify for the task, we re-calibrate the model's prediction probability using this estimated bias.  %Consider the sarcasm detection example again. The sampled random words are likely to contain positive words. Calibrating using prior estimated with random in-domain words thus controls the bias from the positive words in the dataset.

%Motivated by contextual calibration \citep{zhao2021calibrate}, which uses a single pre-defined token (\eg, ``N/A'') to approximate a content-free text for estimating label biases, we use random words sampled from the unlabeled evaluation dataset as content-free text. Consider the sarcasm detection example again. The sampled random words are likely to contain positive words. Calibrating using prior estimated with random in-domain words thus controls the bias from the positive words in the dataset.

More formally, given a dataset, we first construct a bag-of-words $\mathcal{B}$ using the unlabeled texts $\{x_i\}$. Assuming $x_i$'s have average length $L$, we sample $L$ words randomly from $\mathcal{B}$ to form a content-free random text, which captures the word distribution of the dataset domain. However, the random text still remains nearly content-free as it is not grammatically meaningful and potentially contains words from all classes, making it suitable for calibration. We repeat this process $M$ times and average the estimated priors:
\begin{equation}\label{eq:prior}
    \bar P(y|C)=\frac{1}{M}\sum_{j=1}^M P(y|[random\ text]_j,C).
\end{equation}
% We choose $M=20$ as it achieves a good balance between computational efficiency and stability of prior estimation (App.~\ref{sec:sampling_analysis}). 
The model then makes predictions according to the following estimate:
\begin{equation*}
    \hat{y}_i = \mathop{\arg\max}_{y\in\mathcal{L}}\frac{P(y|x_i,C)}{\bar P(y|C)}.
\end{equation*}
where $P(y|x_i,C)$ is the original probability assigned to label $y$ for a particular example $x_i$.

\begin{figure*}[t]
\centering
\includegraphics[width=\textwidth]{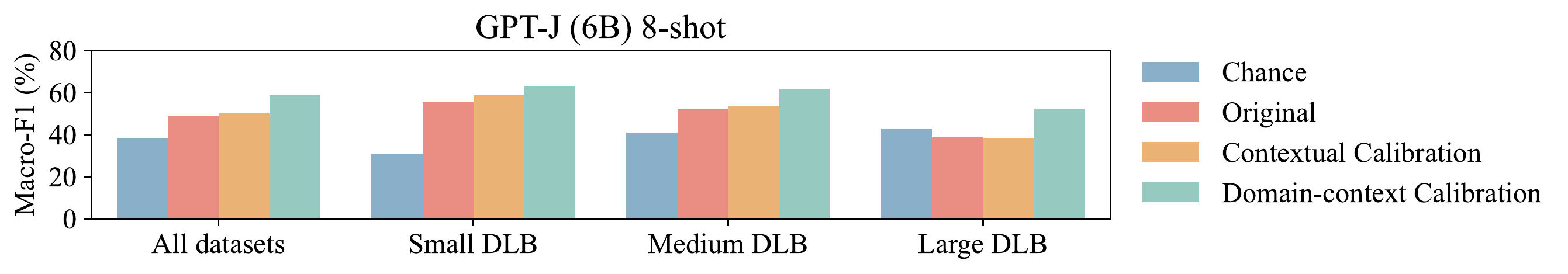}
\includegraphics[width=\textwidth]{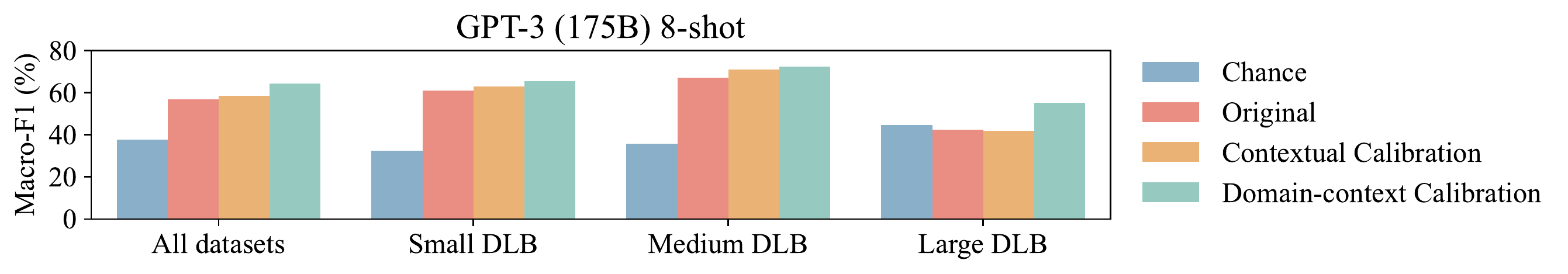}
\caption{The 8-shot ICL performance of GPT-J and GPT-3. We report both the aggregated results on all datasets as well as the average scores on datasets with different levels of \dlb~(DLB). \textbf{Domain-context calibration improves the performance of both models substantially, with larger gains on datasets having larger DLB.}}
\label{fig:main}
\end{figure*}
\begin{figure*}[t]
\centering
\includegraphics[width=0.32\textwidth]{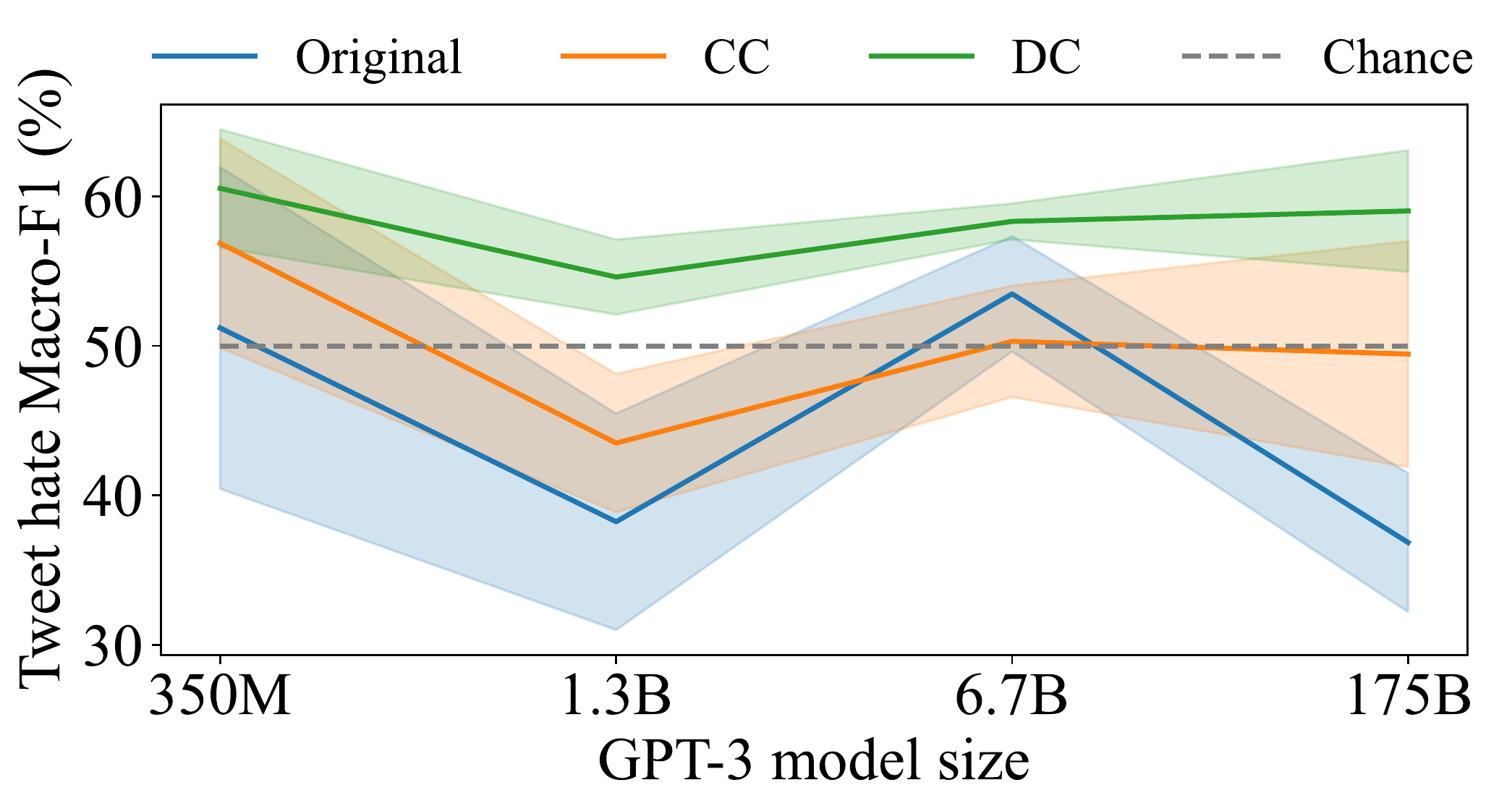}
\includegraphics[width=0.32\textwidth]{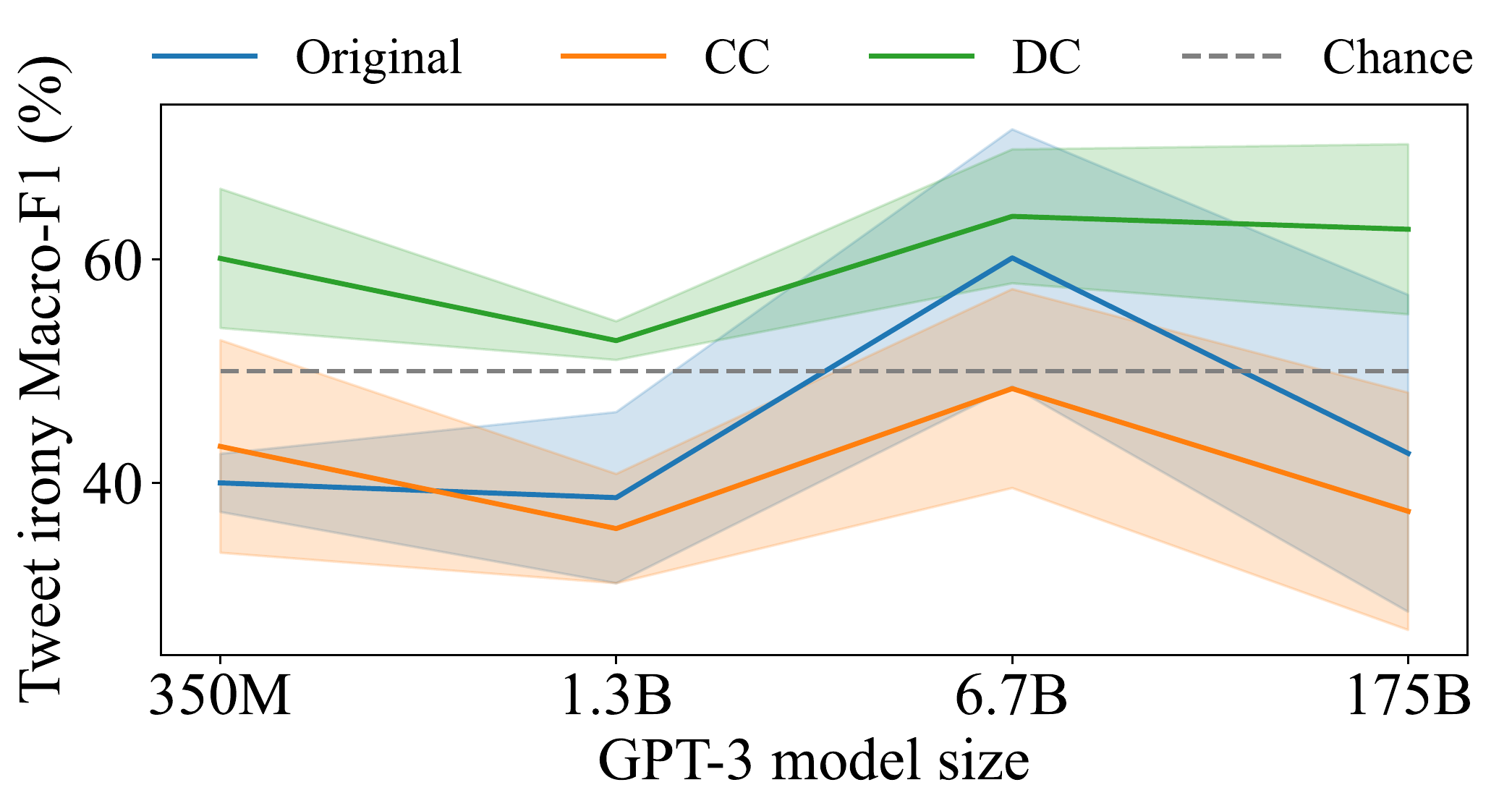}
\includegraphics[width=0.32\textwidth]{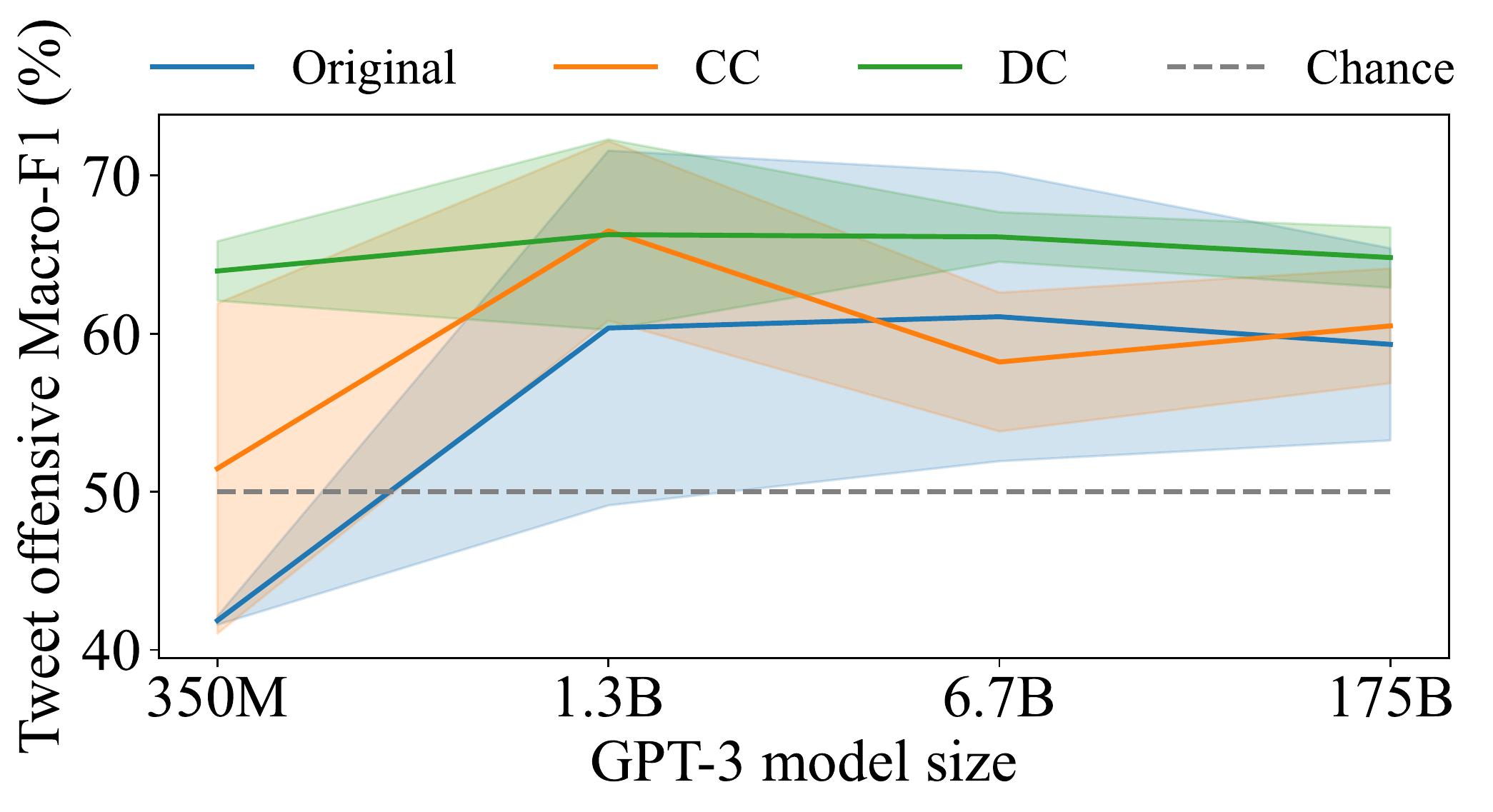}
\caption{The average Macro-F1 ($\pm$ one standard deviation) plots of GPT-3 with five different choices of in-context examples and different model sizes on three datasets with large domain-label bias. \textbf{Scaling up alone does not always increase the performance on tasks with severe domain-label bias. DC boosts the performance of GPT-3 of all different sizes while reducing the variance due to different choices of in-context examples.}}
\label{fig:scaling}
\end{figure*}
%%%%%%%%%%%%%%%%%%%%%%%%%%%%%%%%%%%%%%%%%%%%%%%%%%%%%%%%%%%%%%%%%%%
%%%%%%%%%%%%%%%%%%%%%%%%%%%%%%%%%%%%%%%%%%%%%%%%%%%%%%%%%%%%%%%%%%%
%%%%%%%%%%%%%%%%%%%%%%%%%%%%%%%%%%%%%%%%%%%%%%%%%%%%%%%%%%%%%%%%%%%

\section{Experimental Setup}
\label{sec:setup}
We conduct comprehensive experiments to analyze the effectiveness of our domain-context calibration in mitigating label biases in ICL.% domain-label bias. %We first explain our experimental setups (\S~\ref{sec:setup}). Then, we evaluate DC on a wide range of datasets (\S~\ref{sec:main}). Next, we study the robustness of DC in mitigating domain-label bias under various ICL settings (\S~\ref{sec:robustness}). We conduct systematic ablation studies to analyze DC (\S~\ref{sec:ablation}). Finally, we discuss DC for zero-shot prompting with smaller transformer models (\S~\ref{sec:zero-shot_prompting}). 

% \subsection{Datasets and Experimental Setup}\label{sec:setup}

\paragraph{Datasets} We conduct experiments on 24 text classification datasets that cover a wide range of tasks. Most of these datasets are recently used for studying ICL \citep{zhao2021calibrate, min2022rethinking, lu2021fantastically}. To control the evaluation budget, we use a subset of the 24 datasets for GPT-3 experiments following \citet{min2022rethinking}.  More details can be found in Appendix~\ref{sec:dataset_info}. 

% \yifan{this part can be moved into appendix}
\paragraph{Model and implementation details} We use GPT-J (6B) and GPT-3 (175B) as models in our study. For all experiments, unless stated otherwise, we use $k=8$ examples sampled randomly from the training set to construct the context prompt and evaluate 5 times using different random seeds. Following \citet{min2022rethinking}, we use simple and unified templates for all datasets and do not use any task instructions to keep human engineering at a minimal level. We discuss the effect of task instructions in \S~\ref{sec:robustness}. The templates and label names we used for all datasets can be found in App.~\ref{sec:app_template}. To further control the budget for evaluating with GPT-3, we follow \citet{lu2021fantastically} and sample a subset of size 500 for all datasets whose test sets have sizes exceeding this number. For domain-context calibration, we use the unlabeled test set for sampling random in-domain words and aggregate using $M=20$ random texts (Eq.~\ref{eq:prior}). We discuss the sampling of random words in more detail in Appendix~\ref{sec:sampling_analysis}. We use Open-AI's API for GPT-3 experiments and Tesla V100 GPUs for GPT-J inference.

\paragraph{Evaluation Details} For each model, we compare the performance of domain-context calibration to the following baselines: random performance, uncalibrated performance, and contextual calibration performance. Following prior work, we use the Macro-F1 score as the evaluation metric.
% \begin{table}[t]
% \centering
% \small
% \begin{tabular}{ccllllll}
% \toprule
% {\bf Shot} & {\bf Instr.} & \multicolumn{3}{c}{\bf GPT-J} & \multicolumn{3}{c}{\bf GPT-3} \\
% & & Ori. & CC & DC & Ori. & CC & RC\\
% \midrule
% 8 & \xmark &50.8	&48.4	&\textbf{64.2}	&47.4	&49.8	&\textbf{61.8}\\
% 16 & \xmark &53.7	&45.9	&\textbf{62.7}	&48.0	&49.8	&\textbf{63.1}\\
% 8 & \cmark &47.9	&41.4	&\textbf{64.2}	&47.7	&54.6	&\textbf{68.6}\\
% \bottomrule
% \end{tabular}
% \caption{\label{tab:shots}
% Average Macro-F1 scores on TweetEval-hate, TweetEval-irony and TweetEval-offensive datasets. \textbf{Simply adding more in-context examples or using task instructions does not remove domain-label bias.}
% }
% \end{table
%%%%%%%%%%%%%%%%%%%%%%%%%%%%%%%%%%%%%%%%%%%%%%%%%%%%%%%%%%%%%%%%%%%
%%%%%%%%%%%%%%%%%%%%%%%%%%%%%%%%%%%%%%%%%%%%%%%%%%%%%%%%%%%%%%%%%%%

\section{Experimental Results}\label{sec:main}
We report the average Macro-F1 scores of GPT-J (6B) and GPT-3 (175B) across the entire evaluation suite in Figure~\ref{fig:main}. Furthermore, we stratify our results into three equal-sized subsets according to their levels of domain-label bias.

Our main finding is that \textbf{domain-context calibration generally improves in-context learning, especially on tasks with large domain-label bias}. On all datasets, DC consistently boosts the performance for both models with an average improvement (Macro-F1) of 20\% (GPT-J) and 18\% (GPT-3).\footnote{See Tab.~\ref{tab:full} in App.~\ref{sec:full_main_results} for results on individual datasets.} As Fig.~\ref{fig:main} shows, the performance gain of DC over baselines (original predictions and CC) largely increases when the degree of domain-label bias increases. On the tasks with the largest domain-label bias, DC is the only method that significantly outperforms the random baseline and achieves up to 37\% (GPT-J) and 35\% (GPT-3) performance improvement over other baselines, indicating that DC effectively mitigates domain-label bias.

\subsection{Generalizability}\label{sec:robustness}
Following the finding that DC improves ICL significantly on datasets with large domain-label bias, we analyze the robustness of DC under changes in model scale, number of in-context learning examples, and task instructions (all of which have been shown to improve ICL). We use three datasets that exhibit a high level of domain-label bias for GPT-J.

\paragraph{Scaling up the model} We evaluate GPT-3 models with sizes ranging from 350M to 175B. As Fig.~\ref{fig:scaling} shows, larger models (both when using original prediction or contextualized calibration) do not exhibit better performance on tasks with large domain-label bias. However, DC consistently improves the performance of GPT-3 models of all sizes while reducing the variance due to different choices of in-context examples.
% \yifan{One problem for the experiment writing: the figure/table is quite complex without enough text description. Maybe we can put more experiments in Appendix and describe the experiment setting and design more clearly. Here, better briefly mention baselines and datasets.} 
% \paragraph{Scaling up is not always beneficial on tasks with large domain-label biases} To study the effect of scaling up, we evaluate 4 GPT-3 models with sizes from 350M (ada) to 175B (davinci). As shown in Fig.~\ref{fig:scaling}, there's no clear trend between scaling up and the in-context learning performance, which indicates that 1) scaling up alone is not the ultimate remedy to domain-label bias; 2) the GPT-3 models of different sizes inherit different biases from their pre-training corpus. In all cases, DC boosts the in-context learning performance of GPT-3 while reducing the variance due to different choices of in-context examples.

\paragraph{Adding more in-context examples} In Table~\ref{tab:shots}, we study the effect of adding more in-context examples by evaluating GPT-J and GPT-3 using 0, 8, and 16 in-context examples. For both models, adding more examples does not seem to benefit the model's original and CC performance on tasks with large domain-label bias. However, in all settings, DC gives the best results, and for GPT-3, DC further improves the performance when provided with more in-context examples.
\begin{table}[t]
\centering
\small
\begin{tabular}{cllllll}
\toprule
{\bf Shot}  & \multicolumn{3}{c}{\bf GPT-J} & \multicolumn{3}{c}{\bf GPT-3} \\
& Ori. & CC & DC & Ori. & CC & DC\\
\midrule
0 &56.0	&30.9	&\textbf{65.9}	&51.5	&36.2	&\textbf{61.4}\\
8 &50.8	&48.4	&\textbf{64.2}	&47.4	&49.8	&\textbf{61.8}\\
16 &53.7	&45.9	&\textbf{62.7}	&48.0	&49.8	&\textbf{63.1}\\
\bottomrule
\end{tabular}
\caption{\label{tab:shots}
Average Macro-F1 scores on TweetEval-hate, TweetEval-irony, and TweetEval-offensive. \textbf{Adding more in-context examples does not resolve domain-label bias, while DC always gives the best results.}
}
\end{table}

\paragraph{Task instructions and instruction-tuning} Instruction-tuning and task instructions have been shown to be beneficial to ICL. As shown in Table~\ref{tab:instruction}, for GPT-3, providing task instructions\footnotemark{} improves the performance with DC much more than the original and CC performance, showing that DC enables GPT-3 to make better use of the task instruction. For Instruct-GPT3 (text-davinci-002), adding task instructions largely improves the model's original performance. Still, DC yields significant improvement, while CC actually hurts the model's performance. 

\footnotetext{The instructions we used can be found in App.~\ref{sec:app_template}.}

\begin{table}[!htbp]
\centering
\small
\begin{tabular}{cllllll}
\toprule
{\bf Task}  & \multicolumn{3}{c}{\bf GPT-3} & \multicolumn{3}{c}{\bf Text-davinci-002} \\
{\bf Instruction}& Ori. & CC & DC & Ori. & CC & DC\\
\midrule
\xmark  &47.4	&49.8	&\textbf{61.8}	&58.5	&57.3	&\textbf{71.9}\\
\cmark  &47.7	&54.6	&\textbf{68.6}	&64.1	&56.7	&\textbf{68.1}\\
\bottomrule
\end{tabular}
\caption{\label{tab:instruction}
Average Macro-F1 scores on three tweet datasets. \textbf{DC benefits instruction-tuning models and works in conjunction with task instructions.}
}
\end{table}
%%%%%%%%%%%%%%%%%%%%%%%%%%%%%%%%%%%%%%%%%%%%%%%%%%%%%%%%%%%%%%%%%%%
%%%%%%%%%%%%%%%%%%%%%%%%%%%%%%%%%%%%%%%%%%%%%%%%%%%%%%%%%%%%%%%%%
\subsection{Analysis}\label{sec:ablation}
\begin{figure}[t]
\centering
\includegraphics[width=\columnwidth]{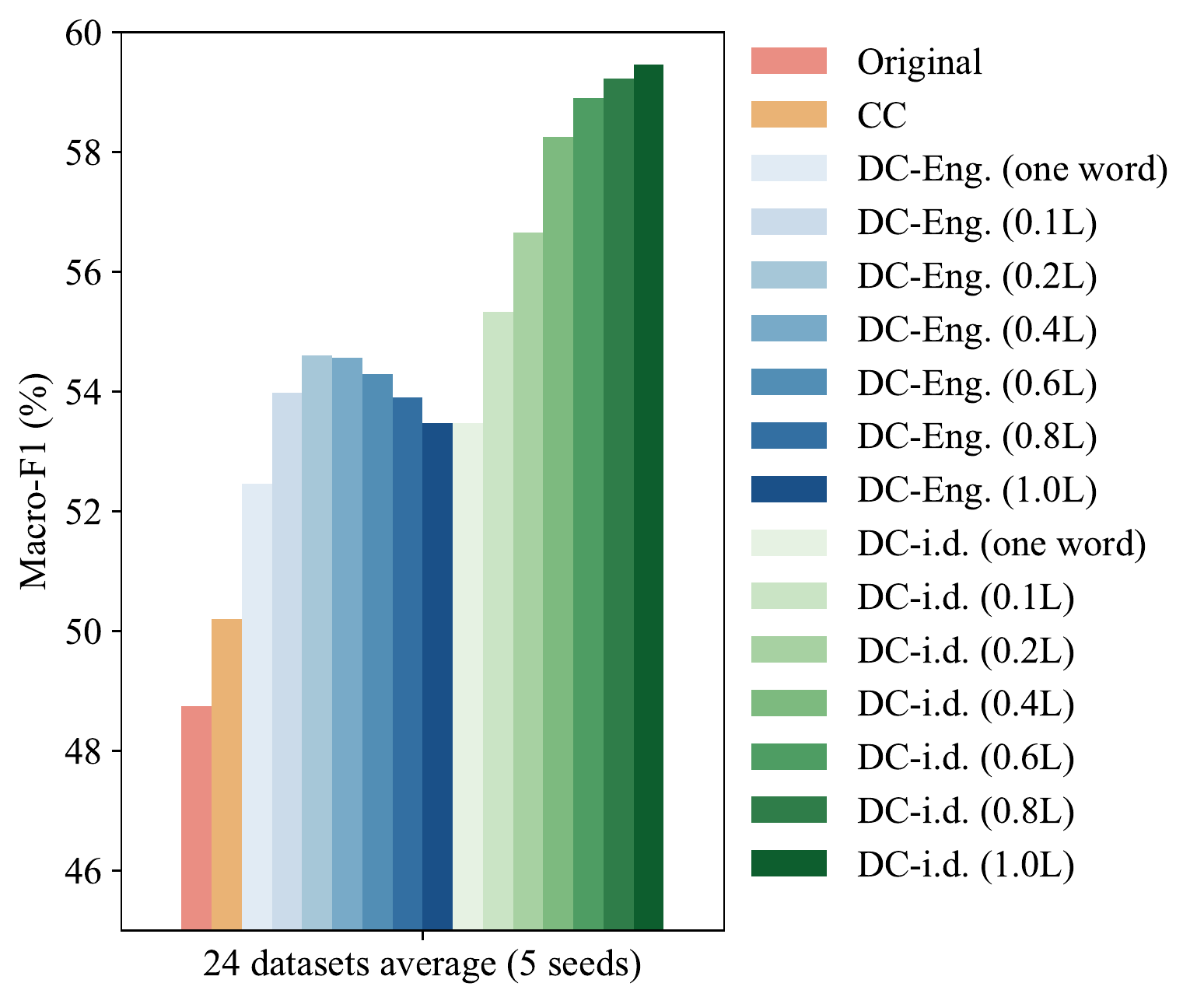}
\caption{Analysis of differences between CC and DC. \textit{DC-Eng.}: DC with random English words. \textit{DC-i.d.}: DC with random in-domain words. The number of random words used in DC is shown in the bracket, where $L$ is the average text length of the evaluation dataset. DC improves CC along three dimensions: 1) replacing ``N/A'' with random words to remove content-free token bias; 2) increasing the number of content-free tokens to estimate the effect of the context prompt more accurately; 3) mitigating domain-label bias by calibrating using in-domain words.
}
\label{fig:ablation_main}
\end{figure}

To understand why DC outperforms CC, we conduct a systematic analysis using GPT-J of three differences between DC and CC: 1) the effect of a predefined content-free token such as ``N/A'' compared to using random words; 2) the length of the random word sequence; 3) the source of random words. Below, we summarize our results from Fig.~\ref{fig:ablation_main}.

% We first replace the pre-defined content-free tokens like "N/A" with random English words to study the effect of the choice of content-free token. Then we increase the number of content-free tokens (i.e., random words) from one to the average text length to study the effect of length. Finally, we compare calibrating with random English words to random in-domain words. 

% We discuss the effect of the number of sampled random texts and the size of the unlabeled task corpus in Appendix~\ref{sec:sampling_analysis}.
\begin{figure*}[t]
\centering 
\includegraphics[width=\textwidth]{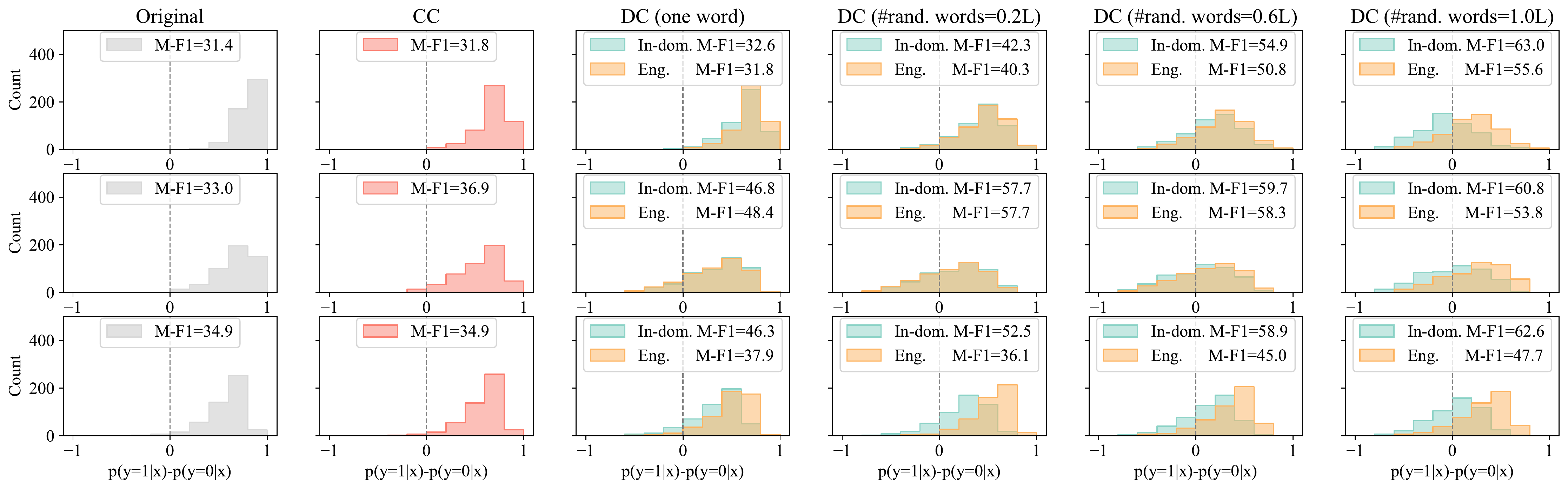}
\caption{The 8-shot GPT-J prediction distributions of three random seeds on TweetEval-hate. Regardless of the choice of in-context examples, GPT-J predicts most inputs as \textit{hate} (gray), even when using CC (red). \textbf{By calibrating using more random in-domain words, the model's bias towards the \textit{hate} class is gradually mitigated (green). Calibrating using more random English words cannot effectively remove domain-label bias (orange).}}
\label{fig:dist_shift}
\end{figure*}
\begin{figure}[t]
\centering
\includegraphics[width=\columnwidth]{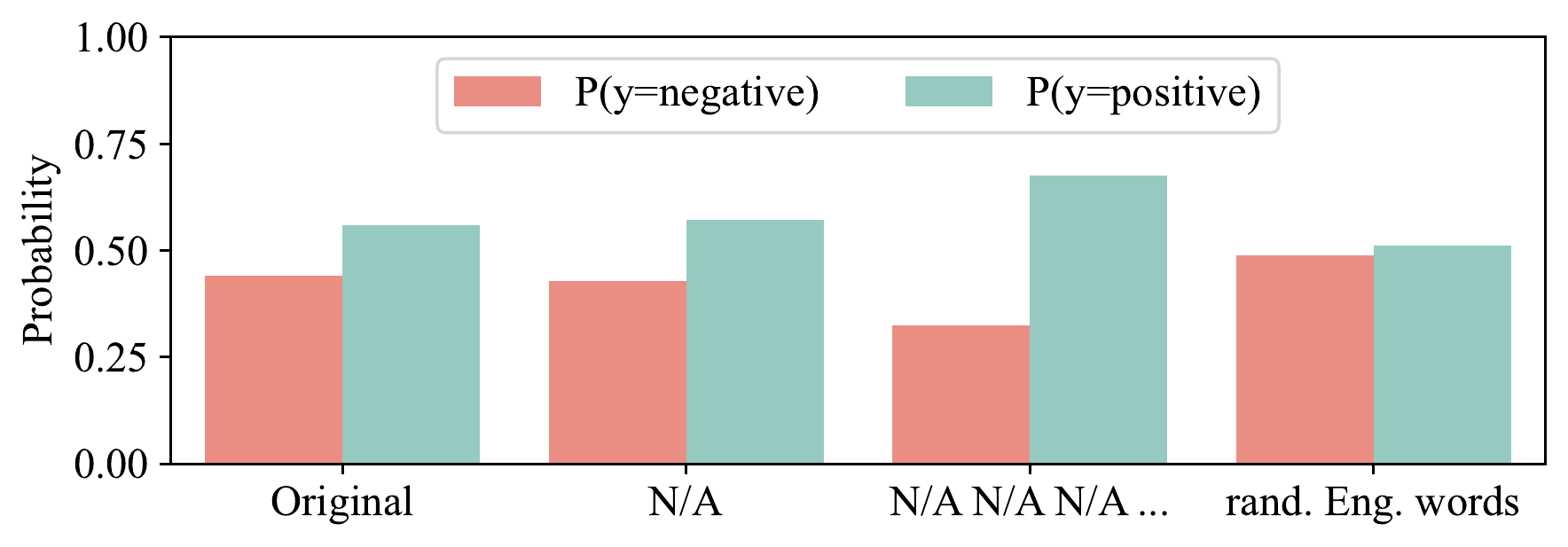}
\caption{We compute the average original and calibrated marginal probabilities of GPT-J on three sentiment datasets (SST-2, CR, and MR; see Table~\ref{tab:datasets} for descriptions). After calibrating with a single content-free token (\eg, ``N/A''), GPT-J is slightly more biased towards the positive class. Such bias increases when we calibrate using a longer sequence of ``N/A''. Calibrating using the same number of random English words does not show such bias, showing that \textbf{predefined content-free tokens also yield label biases.} }
\label{fig:ablation_content_free}
\end{figure}

\paragraph{``Content-free'' tokens can be biased} First, we find that replacing the pre-defined content-free token from CC (\ie, ``N/A'') with a single random English word improves GPT-J's overall performance, indicating that specific content-free tokens may themselves be biased toward particular labels. For example, as shown in Fig.~\ref{fig:ablation_content_free}, on sentiment tasks, calibrating GPT-J using ``N/A'' leads to a systematic bias toward the positive class. Calibrating using random English words to estimate the label bias avoids this problem.

% \yifan{It's a bit distracting to mention multiple figures/tables in one paragraph. Here, when you refer Figure 10, readers need to read Figure 10 as well as its title. When they get back here, they may not remember the point you're giving here. Try to avoid that or using two paragraphs for them.} 

\begin{figure}[t]
\centering
\includegraphics[width=\columnwidth]{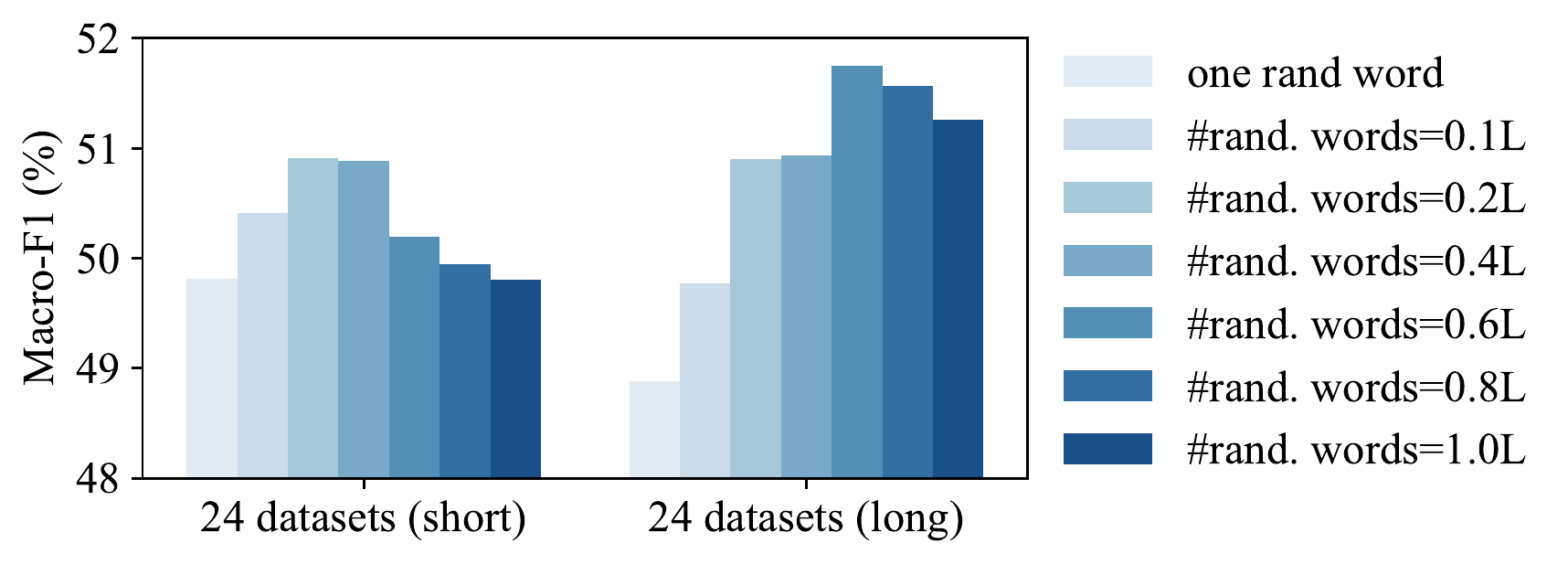}
\caption{The GPT-J 8-shot performance (across 5 random seeds) of calibrating using different numbers of random English words on the 10\% longest and shortest texts of all 24 datasets. $L$ is the average text length of the evaluation dataset. \textbf{Longer (shorter) texts prefer calibrating with longer (shorter) random texts.}}
\label{fig:ablation_length}
\end{figure}

\paragraph{Calibrating with random texts of the average input length is beneficial}
As shown in Fig.~\ref{fig:ablation_main}, when calibrating using random English words, increasing the number of words improves performance. Intuitively, using random texts of the average input length for calibration gives a more precise estimate of the effect of the context prompt. To test this, we select the longest and shortest 10\% samples of all 24 datasets to construct a dataset with long and short inputs for a task. Then, we test the calibration performance using random English words of different lengths.\footnote{We use random English words rather than random in-domain words to better study the effect of the length.} As shown in Fig.~\ref{fig:ablation_length}, longer (shorter) texts prefer longer (shorter) random texts as calibration sequences to estimate the label bias.

\paragraph{Calibrating using random in-domain words removes domain-label bias} Finally, calibrating using random in-domain words yields a large improvement over calibrating using random English words. In Fig.~\ref{fig:dist_shift}, we plot the prediction distributions of {GPT-J} on TweetEval-hate after calibrating with both random English and in-domain words of various lengths. We see that when only calibrating using a few in-domain words, the word distribution of the dataset is not well-captured, and thus the domain-label bias is not effectively removed. When calibrating using more in-domain words, the prediction becomes more balanced. In contrast, after calibrating using more random English words, the model is still biased towards predicting label \textit{hate}. Interestingly, we notice that the more DC mitigates the domain-label bias, the more task performance increases.

% \begin{table}[t]
% \centering
% \small
% \begin{tabular}{clll}
% \toprule
% {\bf Task}& \textbf{Original} & \textbf{CC} & \textbf{RC}\\
% \midrule
% SST-2 & $74.4\pm 11.4$ & $82.4\pm 5.0$ & $82.8\pm 5.4$\\
% AGNews &$56.0\pm 2.2$ & $63.9\pm 5.0$ & $71.4\pm 0.7$\\
% Tweet-hate &$33.9\pm 3.7$ & $41.7\pm 2.3$ & $61.1\pm 4.7$\\
% Tweet-irony &$40.3\pm 18.5$ & $42.0\pm 2.3$ & $55.6\pm 12.5$\\
% \bottomrule
% \end{tabular}
% \caption{\label{tab:smaller}
% The average Macro-F1 scores of zero-shot prompting with RoBERTa-large using 3 different templates. \textbf{DC benefits prompting with smaller LLMs.}
% }
% \end{table}

%%%%%%%%%%%%%%%%%%%%%%%%%%%%%%%%%%%%%%%%%%%%%%%%%%%%%%%%%%%%%%%%%%%
%%%%%%%%%%%%%%%%%%%%%%%%%%%%%%%%%%%%%%%%%%%%%%%%%%%%%%%%%%%%%%%%%%%
\subsection{Zero-shot Prompting}\label{sec:zero-shot_prompting}
Smaller LLMs pre-trained using masked language modeling can also be efficiently adapted to unseen tasks by reformulating the task into a cloze problem using a natural language prompt (\ie, zero-shot prompting; \citealp{schick2020exploiting}). To demonstrate that DC can be used even with smaller models, we evaluate the zero-shot prompting ability of RoBERTa-large \citep{liu2019roberta} when it is calibrated using DC.\footnote{Implementation details can be found in App.~\ref{sec:zero-shot}} We report our results in Tab.~\ref{tab:zeroshot_prompting} (App.~\ref{sec:zero-shot}) and find that across the same 24 datasets, DC achieves a significant performance gain of 26\%. Similar to ICL with GPT models, label biases affect RoBERTa's zero-shot prompting priors. However, DC effectively mitigates these biases, leading to significant performance improvements.

%%%%%%%%%%%%%%%%%%%%%%%%%%%%%%%%%%%%%%%%%%%%%%%%%%%%%%%%%%%%%%%%%%%
%%%%%%%%%%%%%%%%%%%%%%%%%%%%%%%%%%%%%%%%%%%%%%%%%%%%%%%%%%%%%%%%%%%
%%%%%%%%%%%%%%%%%%%%%%%%%%%%%%%%%%%%%%%%%%%%%%%%%%%%%%%%%%%%%%%%%%%
\section{Discussion}\label{sec:discussion}

\begin{figure}[t]
\centering
\includegraphics[width=\columnwidth]{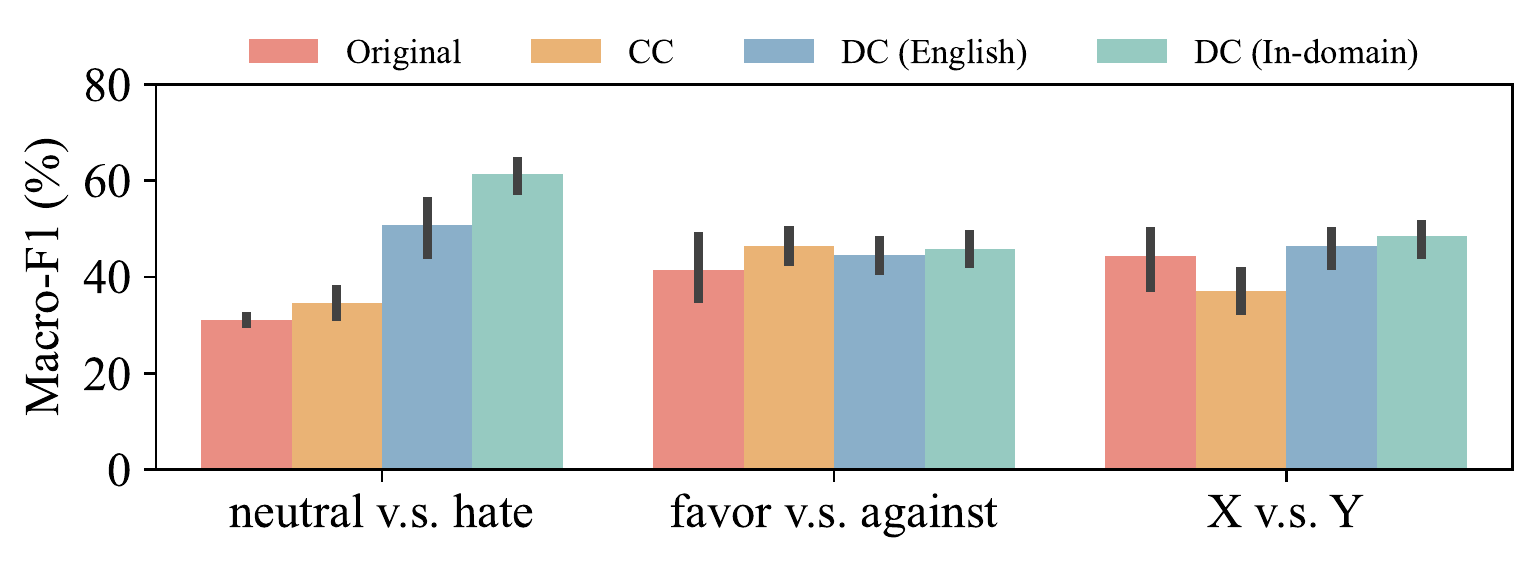}
\caption{The Macro-F1 scores of GPT-J (8-shots) on TweetEval-hate with different label names. DC (English): DC with random English words; DC (In-domain): DC with random in-domain words. \textbf{Using less task-relevant label names mitigates domain-label bias but limits the model's performance.}}
\label{fig:ablation_verbalizer}
\end{figure}

\paragraph{Label name selection as label bias mitigation} Our results outline how LLMs can be biased to certain label names for different tasks. Intuitively, because the task is formulated as \textit{generating} the label name given an example, the mechanism elicits the model's prior knowledge about the task. To better understand whether domain-label bias could be mitigated through more careful label name selection, we test GPT-J with three different pairs of label names on TweetEval-hate: 1) \textit{neutral} v.s. \textit{hate}, which is the most task-relevant set of label names, but introduces severe domain-label bias; 2) \textit{favor} v.s. \textit{against}, a pair of less task-relevant antonyms used by \citet{min2022rethinking}; 3) \textit{X} v.s. \textit{Y}, which are meaningless placeholders. 

As shown in Fig.~\ref{fig:ablation_verbalizer}, calibrating using random English words or in-domain words makes little difference when choosing (\textit{X}, \textit{Y}) or (\textit{favor}, \textit{against}) as the label names, showing that they do not introduce domain-label bias. However, although GPT-J is able to achieve better original and CC performance on these label names, (\textit{neutral}, \textit{hate}) yields the best performance after removing domain-label bias using DC. Thus, with proper calibration, using the most task-indicative words as label names is likely to be the best option. Surprisingly, the manually picked label names (\textit{favor}, \textit{against}) under-perform the meaningless ones (\textit{X}, \textit{Y}) after applying DC, {hinting that human-plausible label names are not necessarily good label names for LLMs. }
% \antoine{why does this result emphasize this importance?}
% emphasizing the importance of choosing task-relevant words as the label names in in-context learning. 

% \paragraph{When should we consider using RC?} As discussed in \S~\ref{sec:main}, the largest improvement of DC over CC and the original in-context learning results happens when the datasets suffer from large domain-label bias. To illustrate this in a more detailed view, we plot the Macro-F1 score gains of DC over CC (GPT-J) on all 24 datasets with the corresponding domain-label biases measured using \eqref{eq:domain_label} in Fig.~\ref{fig:correlation}. We see that the larger the measure, the greater the benefit of applying DC tends to be. 

\paragraph{Selecting datasets for ICL analysis} The varying levels of domain-label bias in our studied datasets suggest a large variance in how ICL will perform on different datasets. Consequently, macro-averaging the performance on datasets with differing levels of label biases potentially obfuscates diverse results among different tasks. Our work encourages future studies to select datasets carefully to cover varying degrees of label bias among reported results, and to perform fine-grained analysis of individual datasets when studying ICL performance.

% \paragraph{Choosing datasets for analyzing ICL} The varying levels of domain-label bias in our studied datasets suggest a large variance in how ICL will perform on different datasets. Consequently, macro-averaging the performance on datasets with differing levels of label biases potentially obfuscates diverse results among different tasks. For example, analyzing a dataset where the model consistently predicts one label regardless of the choice of in-context examples hinders developing a good understanding of the importance of in-context examples. Thus our work motivates 1) a rigorous design on dataset selection and formulation to account for confounders and 2) a fine-grained analysis of individual datasets to study ICL more effectively. 

\paragraph{Alternate causes of domain label bias} When evaluating models for real-world applications such as hate speech detection, we usually use hard examples (\eg, non-hateful, but ``hateful-looking'' examples) to check the robustness of the model. However, LLMs trained on natural corpora are likely to be susceptible to adversarial word-level features (LLMs use word associations learned from pre-training to perform ICL). To some degree, adversarial examples could also be a source of large domain-label bias on many datasets. %The existence of such bias indicates that LLMs use word associations learned from pre-training to perform ICL.
% \subsection{Analysis}
% % \paragraph{Random text calibration also works for instruct-GPT} aaa aaa aaa aaa aaa aaa aaa aaa aaa aaa aaa aaa aaa aaa aaa aaa aaa aaa aaa aaa aaa aaa aaa aaa aaa aaa aaa aaa aaa aaa aaa aaa aaa aaa aaa aaa aaa aaa aaa aaa aaa aaa aaa aaa aaa aaa aaa aaa aaa aaa aaa aaa aaa aaa aaa aaa aaa.

% \paragraph{Simply adding more few-shot examples or task instruction do not solve the problem}
% aaa aaa aaa aaa aaa aaa aaa aaa aaa aaa aaa aaa aaa aaa aaa aaa aaa aaa aaa aaa aaa aaa aaa aaa aaa aaa aaa aaa aaa aaa aaa aaa aaa aaa aaa aaa aaa aaa aaa aaa aaa aaa aaa aaa aaa aaa aaa aaa aaa aaa aaa aaa aaa aaa aaa aaa aaa.

% \paragraph{Use task irrelevant class names relieves domain-label bias but hurts model's performance}
% aaa aaa aaa aaa aaa aaa aaa aaa aaa aaa aaa aaa aaa aaa aaa aaa aaa aaa aaa aaa aaa aaa aaa aaa aaa aaa aaa aaa aaa aaa aaa aaa aaa aaa aaa aaa aaa aaa aaa aaa aaa aaa aaa aaa aaa aaa aaa aaa aaa aaa aaa aaa aaa aaa aaa aaa aaa.
% \section{Discussion}
% \paragraph{How important are the label names?}

%%%%%%%%%%%%%%%%%%%%%%%%%%%%%%%%%%%%%%%%%%%%%%%%%%%%%%%%%%%%%%%%%%%
%%%%%%%%%%%%%%%%%%%%%%%%%%%%%%%%%%%%%%%%%%%%%%%%%%%%%%%%%%%%%%%%%%%
%%%%%%%%%%%%%%%%%%%%%%%%%%%%%%%%%%%%%%%%%%%%%%%%%%%%%%%%%%%%%%%%%%%
\section{Related Work}
In-context learning (ICL) is the standard paradigm for adapting LLMs \citep{chowdhery2022palm, wei2022chain, zhang2022opt}. Many recent works focus on understanding its mechanism to improve adaptation performance. For example, \citet{lu2021fantastically} showed that ICL is sensitive to the order of in-context examples. \citet{razeghi2022impact} demonstrated that the ICL performance on numerical reasoning tasks is strongly correlated with the pre-training term frequencies. \citet{liu2021makes} found that using examples semantically close to the input texts is beneficial.  \citet{min2022rethinking} showed that for classification tasks, the input-label pairing format plays the most crucial role in ICL. \citet{sorensen2022information} found that the structure of the prompt also significantly affects ICL performance, and that better prompts could be selected based on mutual information between the prompts and the model's output. Complementary to these works, we comprehensively study the label bias problem in ICL. The existence of domain-label bias indicates that ICL is largely affected by the word-level associations LLMs learn during pre-training.%We also shed light on future work to consider domain-label bias when analyzing ICL.
% this line of analysis work and demonstrate that the distribution of the downstream dataset can lead to severe biases, causing the model to barely outperform the random baseline on a wide range of tasks. \TODO{Rewrite}

Other recent works discuss the bias problem in ICL. \citet{zhao2021calibrate} proposed contextual calibration to mitigate three types of biases in ICL: the majority bias, recency bias, and common token bias. \citet{holtzman2021surface} focused on the zero-shot setting and found that different surface forms of the answer can compete for probability mass given a question, leading to a bias when predicting with a single label name for each class. In contrast to these works, which consider a specific type of bias, we propose a typology of label biases and propose domain-context calibration that handles all biases in our typology. %\antoine{}

Finally, the ability of the largest models, such as PaLM \citep{chowdhery2022palm}, to perform in-context learning flipped labels or semantically-unrelated label settings \citep{wei2023larger} is very relevant to this work. As the largest models tend to have emergent abilities, it would be interesting to test how vulnerable these models are to label biases (especially domain-label bias) and how domain-context calibration would help. Unfortunately, we currently do not have access to them (\eg, PaLM; Flan-PaLM, \citealp{chung2022scaling}). Nevertheless, the similar behavior of PaLM and Instruct-GPT (as shown in \citealp{wei2023larger}) and the fact that Instruct-GPT also suffers from domain-label bias (Table \ref{tab:instruction}) indicate that these more capable models may still be susceptible to label biases. %Also, how scaling up or instruction-tuning would relieve label biases is an interesting direction to explore.

% \citet{han2022prototypical} viewed the unstableness of ICL as the result of prediction distribution shifts and proposed a clustering-based approach to improve the prediction. 
% A particularly similar work to ours is \citep{zhao2021calibrate}. They proposed contextual calibration to mitigate three types of biases in ICL: the majority bias, recency bias, and common token bias, which are related to the ordering of in-context examples and the specific choice of label names. Adding to their work, we show that ICL suffers from task-relevant biases caused by input word distributions. Additionally, we identify two drawbacks of their calibration setting, which our proposed calibration method can successfully resolve. \TODO{Add comparision with prototypical calibration \citep{han2022prototypical} and surface form competition paper \citep{holtzman2021surface}}
%%%%%%%%%%%%%%%%%%%%%%%%%%%%%%%%%%%%%%%%%%%%%%%%%%%%%%%%%%%%%%%%%%%
%%%%%%%%%%%%%%%%%%%%%%%%%%%%%%%%%%%%%%%%%%%%%%%%%%%%%%%%%%%%%%%%%%%
%%%%%%%%%%%%%%%%%%%%%%%%%%%%%%%%%%%%%%%%%%%%%%%%%%%%%%%%%%%%%%%%%%%

\section{Conclusion}
In this work, we define a typology of label biases that affect in-context learning (ICL). We categorize existing findings of label biases into two types: vanilla-label bias and context-label bias, and identify a new type of bias, domain-label bias, that significantly influences ICL performance. To mitigate these label biases, we propose domain-context calibration, which significantly improves ICL performance on a wide range of tasks, particularly on datasets with large domain-label bias. %\bluetext{
%From a broader view, our work provides a comprehensive framework for analyzing label biases in ICL. 
The various levels of domain-label bias in different datasets also suggest that when analyzing ICL, we need to select datasets with diverse types of label biases and report stratified results that acknowledge this diversity beyond single aggregated scores. %over a collection of datasets without careful selection.}

% \TODO{Needs to rewrite from the broader view} We show that the word distribution of the target dataset can severely bias the model towards a particular class in few-shot in-context learning. Such domain-label bias causes LLMs with different sizes and pre-training methods to barely perform the random baseline on a wide range of tasks. We propose a simple calibration method that largely mitigates the domain-label bias and further fixes two drawbacks of contextual calibration. 

% From a broader view, our work provide a more complete picture of the bias sources in in-context learning. The existence of domain-label bias also suggests that: when analyzing in-context learning, we need to go beyond focusing on a particular type of dataset, e.g., topic and sentiment classification, or making conclusions based aggregated scores over datasets with different types. To get a more fine-grained understanding of in-context learning, we should study these datasets individually or group them by their types.


\section*{Limitations}
\paragraph{Data and Task Limitation}
In this work, we analyze domain-label bias and apply our domain-context calibration to English. We leave analysis and mitigation methods for multilingual tasks to future works. In experiments, we discuss calibration on classification tasks. The effect of domain-label bias could exist differently for open-end tasks like text generation. Our analysis of domain-label bias also emphasizes more on the word-level bias. Other types of biases associated with a domain, such as topics and genders, may also impact model prediction. We leave the diverse analysis to future works. Due to budget limitations, we conduct experiments on a subset of the 24 reported datasets for GPT-3. One can evaluate all 24 datasets to get a complete picture with enough budget.

\paragraph{Model Limitation}
For large language models, we only focus on the GPT models and only select RoBERTa as the small-scale language model in experiments. Future work could consider expanding to other model types, such as PaLM for large models and DeBERTa for small models. Access to the OpenAI API for GPT-3 is also necessary for parts of our experiments. Future work can consider experimenting with open-source LLMs like the OPT-175B or BLOOM-176B model.

\section*{Ethics Statement}
Our work focuses on analyzing the general label bias problem of the in-context learning ability of LLMs and improving their performance with a tuning-free method, which involves no large neural model pre-training, re-training, or fine-tuning. As we only use LLMs for inference, developing and applying our approach requires only minimal computational resources compared to methods that require dataset-specific model fine-tuning or engineering. We do not anticipate signiﬁcant ethical issues introduced by our approach, as we use only off-the-shelf LLMs, and the datasets involved are all publicly available text classification datasets. The discussion of biases in our work is general and not specific to any real word context. Still, our analysis and typology of the label biases of ICL may motivate future work to analyze the bias problem of ICL and LLMs in areas with larger social impacts, such as healthcare or legal scenarios.

\section*{Acknowledgements} We thank Silin Gao, Syrielle Montariol, and Gail Weiss for helpful feedback on this paper. We also gratefully acknowledge the support of Innosuisse under PFFS-21-29, the EPFL Science Seed Fund, the EPFL Center for Imaging, Sony Group Corporation, and the Allen Institute for AI.

%Indeed, we highlight how the pretraining corpora of large language models might be sources of critical biases in downstream tasks - here, through what we denote vanilla-label and domain-label bias. Our work aims at mitigating these biases, contributing to improving the LLM's ability to generalize from a few examples.

\bibliography{acl2023}
\bibliographystyle{acl_natbib}

\appendix
\newpage
\section{Domain-label Bias of All Datasets}\label{sec:domain-label_bias_all}
\begin{figure*}[t]
\centering
\includegraphics[width=\textwidth]{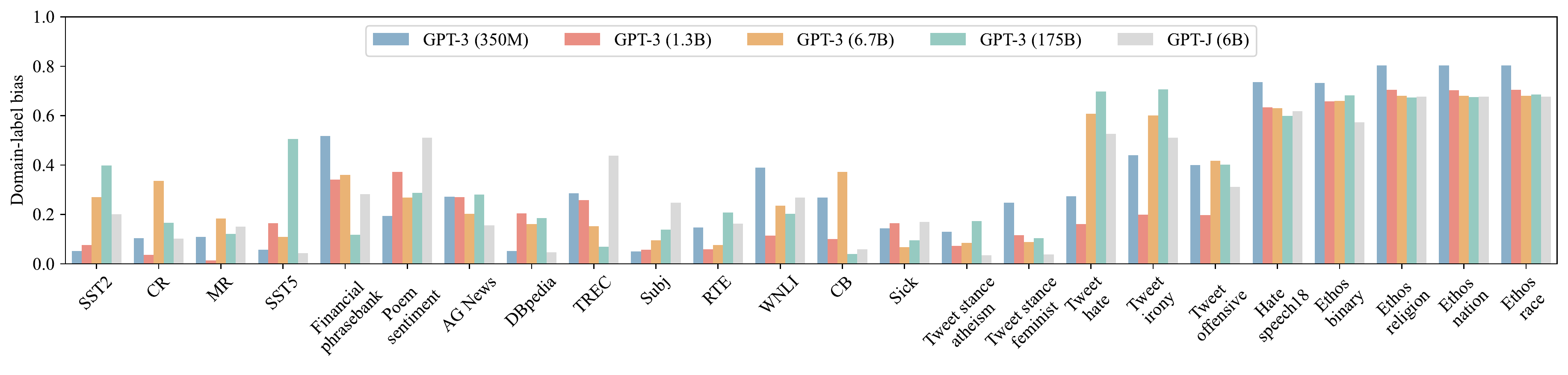}
\includegraphics[width=\textwidth]{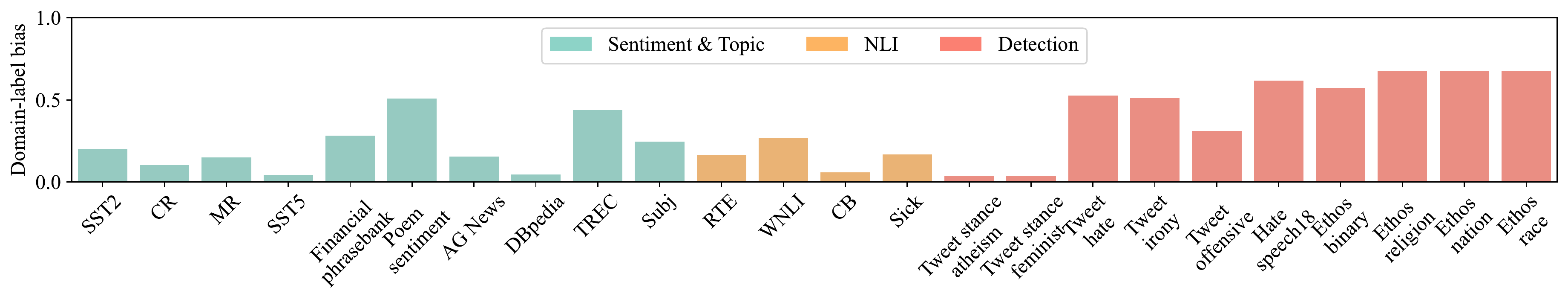}
\caption{Top: Domain-label bias of different models on all evaluation datasets. Bottom: Domain-label bias of GPT-J on all evaluation datasets categorized based on the task types.}
\label{fig:domain-label_dataset}
\end{figure*}

We compute and illustrate the domain-label bias of all datasets we used with different LLMs in Fig.~\ref{fig:domain-label_dataset}. Different datasets exhibit different levels of domain-label bias. Regarding task types, The detection tasks (red) show the largest domain-label bias, while the NLI tasks (orange) have the least. On sentiment and topic tasks, the domain-label bias is mostly small but can vary depending on the domain of the dataset. For example, for sentiment classification datasets, movie review datasets like SST-2 have relatively small domain-label bias. While financial statements and poem sentiment, whose texts are from rare domains, have much larger biases. We discuss the model dependency of domain-label bias in the next section.

\section{Correlation of Domain-label Bias Estimated with Different LLMs}\label{sec:correlation}
\begin{figure}[ht]
\centering
\includegraphics[width=\columnwidth]{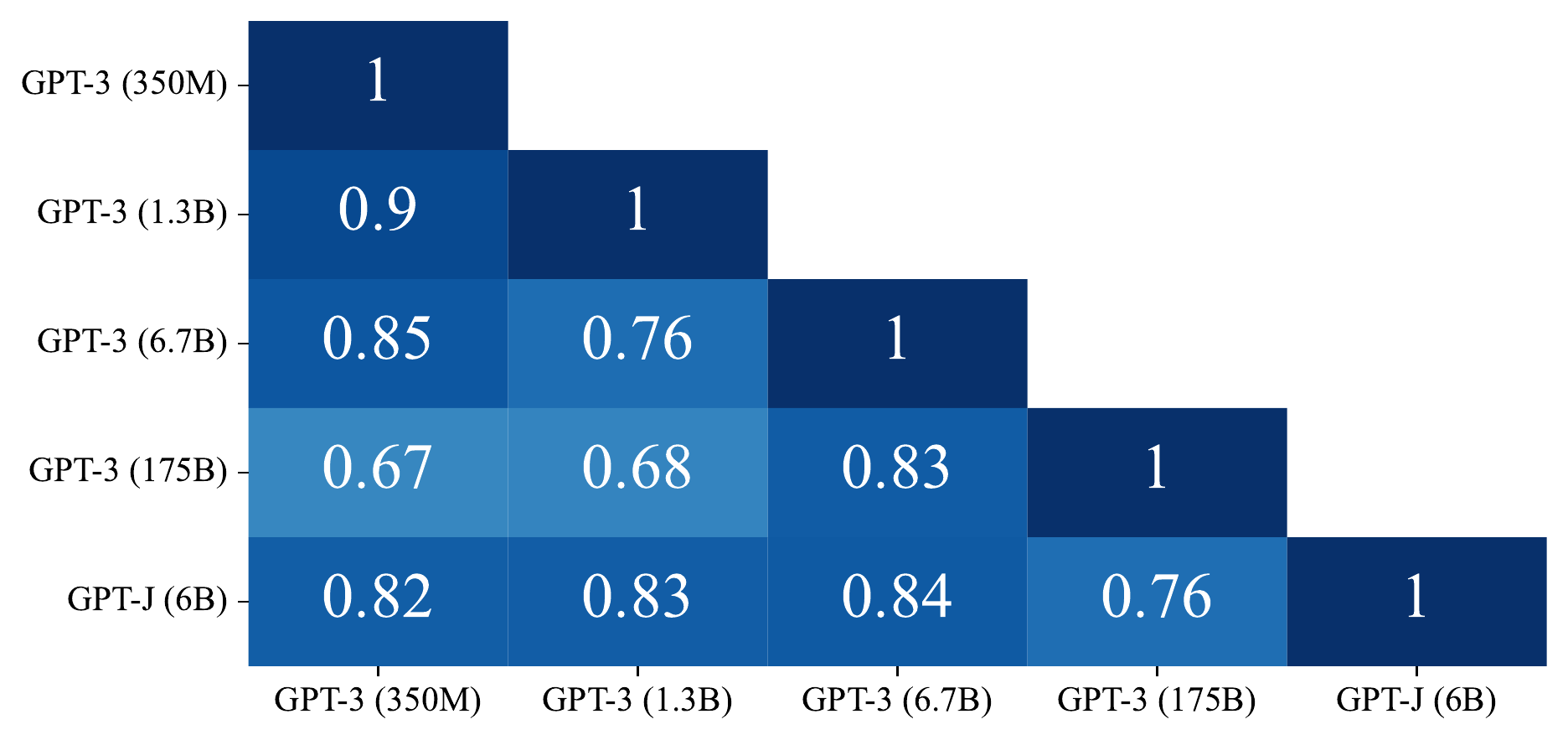}
\caption{Although \dlb~is model-dependent, we observe a high correlation of \dlb~between LLMs on the evaluation datasets.}
\label{fig:model_correlation}
\end{figure}

We compute the correlation of domain-label bias (defined by eq.~\eqref{eq:domain_label}) computed with 5 different models on all 24 evaluation datasets. We use GPT-3 Ada (350M), GPT-3 Babbage (1.3B), GPT-3 curie (6.7B), GPT-3 DaVinci (175B), and GPT-J (6B). We show the correlation plot in Figure \ref{fig:model_correlation}. Although domain-label bias is model-dependent by definition, the biases computed by different models are highly correlated. 
\section{Full Dataset Information} \label{sec:dataset_info}
We use 24 datasets falling into three categories: sentiment and topic classification, NLI, and Detection. Most of the used datasets are from existing works. W\citep{min2022rethinking, lu2021fantastically, zhao2021calibrate}e added a few more detection datasets for better studying the domain-label bias as they tend to show the largest domain-label bias.  We use the HuggingFace venison \citet{lhoest2021datasets} of all datasets and use the test set, if available, for evaluation. Otherwise, we use the development set. We summarize the full dataset information in Table~\ref{tab:datasets}. 

\section{Full Few-shot Results} \label{sec:full_main_results}
\begin{figure}[ht]
\centering
\includegraphics[width=\columnwidth]{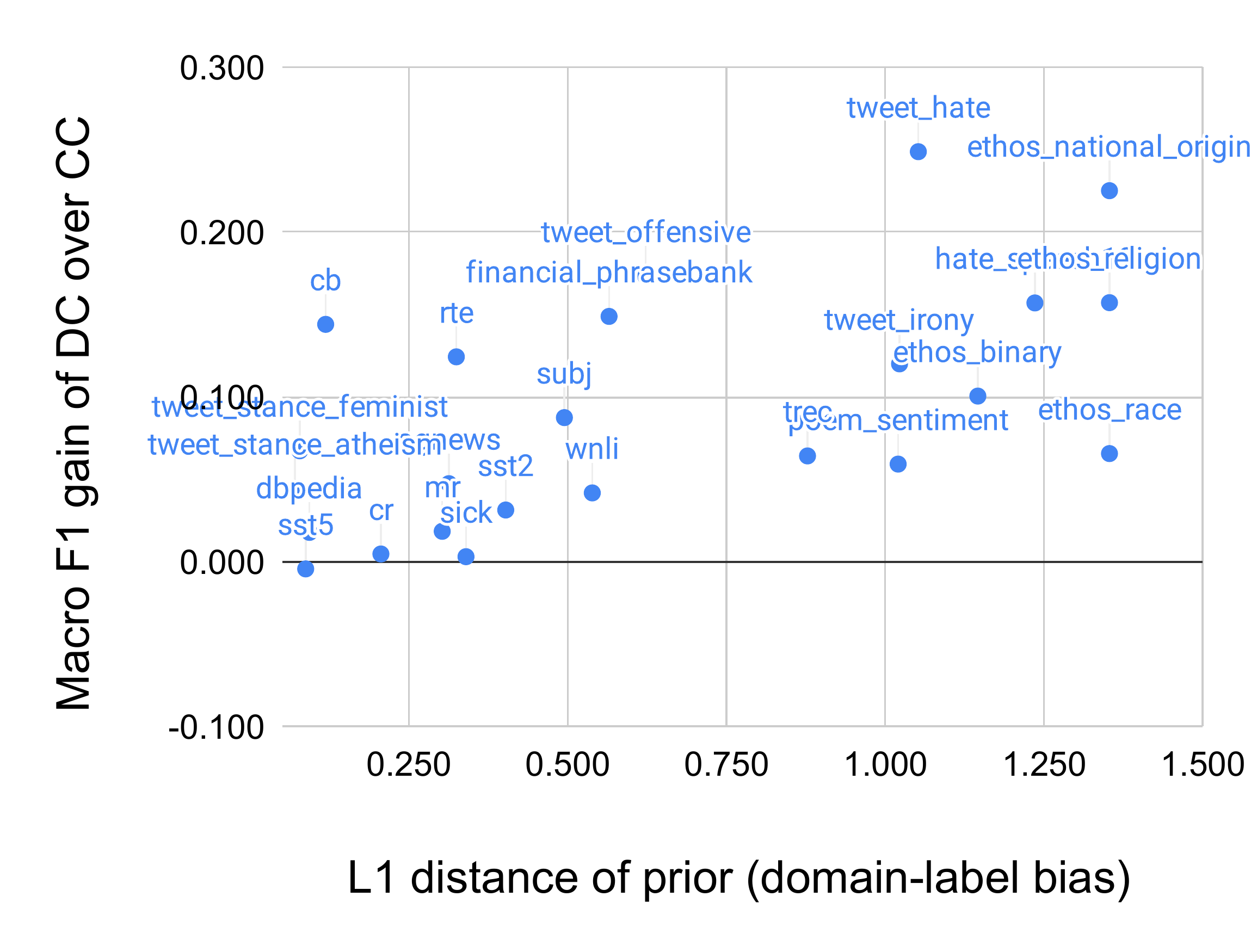}
\caption{On tasks with more severe domain-label bias, we can observe more performance gain of DC over CC.}
\label{fig:correlation}
\end{figure}

We report the full 8-shot results on individual datasets with the standard deviations (5 random seeds) in Table~\ref{tab:full}. We further show the few-shot performance gain (GPT-J) of DC over CC on all datasets in Figure \ref{fig:correlation}.
\section{Templates and Task Instructions} \label{sec:app_template}
We show the templates and label names for all datasets in Table~\ref{tab:templates}. The task instructions used in Table~\ref{tab:instruction} are illustrated in Table~\ref{tab:task_instructions}. We always use exactly one word for every label name. To avoid label names being tokenized into subwords, we always use lower-cased label names except for tasks answering with True or False.

\section{Sampling Analysis} \label{sec:sampling_analysis}
In this section, we analyze two factors related to the random word sampling process involved in DC: 1) the number of random texts to use for estimating the prior ($M$ in eq.~\eqref{eq:prior}), and 2) the size of the unlabeled dataset to sample random in-domain words from. We conduct experiments on two small-domain-label-bias datasets (SST-2 and AG News) and two large-domain-label-bias datasets (TweetEval-hate and TweetEval-irony).

\begin{figure}[h]
\centering
\includegraphics[width=\columnwidth]{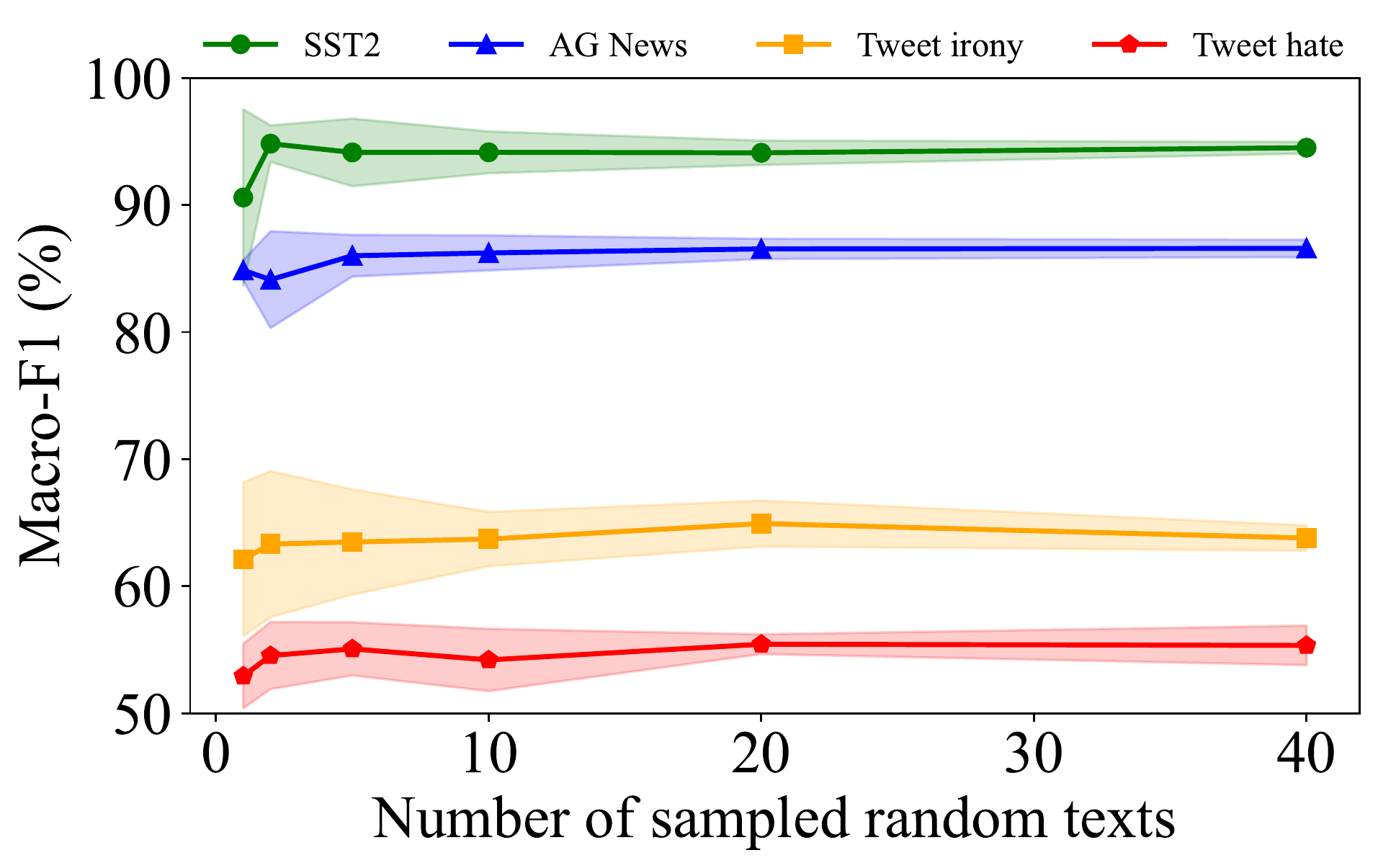}
\caption{The few-shot performance of GPT-J after applying domain-context calibration with different numbers of sampled random texts for estimating the model's prior. For each number, we sample random texts 5 times with different random seeds to show the stability of the sampling process. Using 20 random texts already provides a stable estimate of label biases.}
\label{fig:sampling_sum}
\end{figure}

\paragraph{How many random texts should we sample?}
First, we sample different numbers of random texts $M$ for estimating the model's prior as in eq.~\eqref{eq:prior}. As shown in Figure \ref{fig:sampling_sum}, DC is able to achieve a good estimate with a relatively small number of sampling. We choose $M=20$ as it achieves a good balance between computational efficiency and stability of prior estimation.

\paragraph{How large should the unlabeled task corpus be?}
In the main experiments, we use the whole unlabeled test set to construct a bag-of-words and sample random words from it. Here, we study the effect of the unlabeled dataset set size on the performance of DC. As shown in Figure \ref{fig:sampling_sum}, DC is able to achieve a good estimate with 50 unlabeled texts from the dataset.
\begin{figure}[ht]
\centering
\includegraphics[width=\columnwidth]{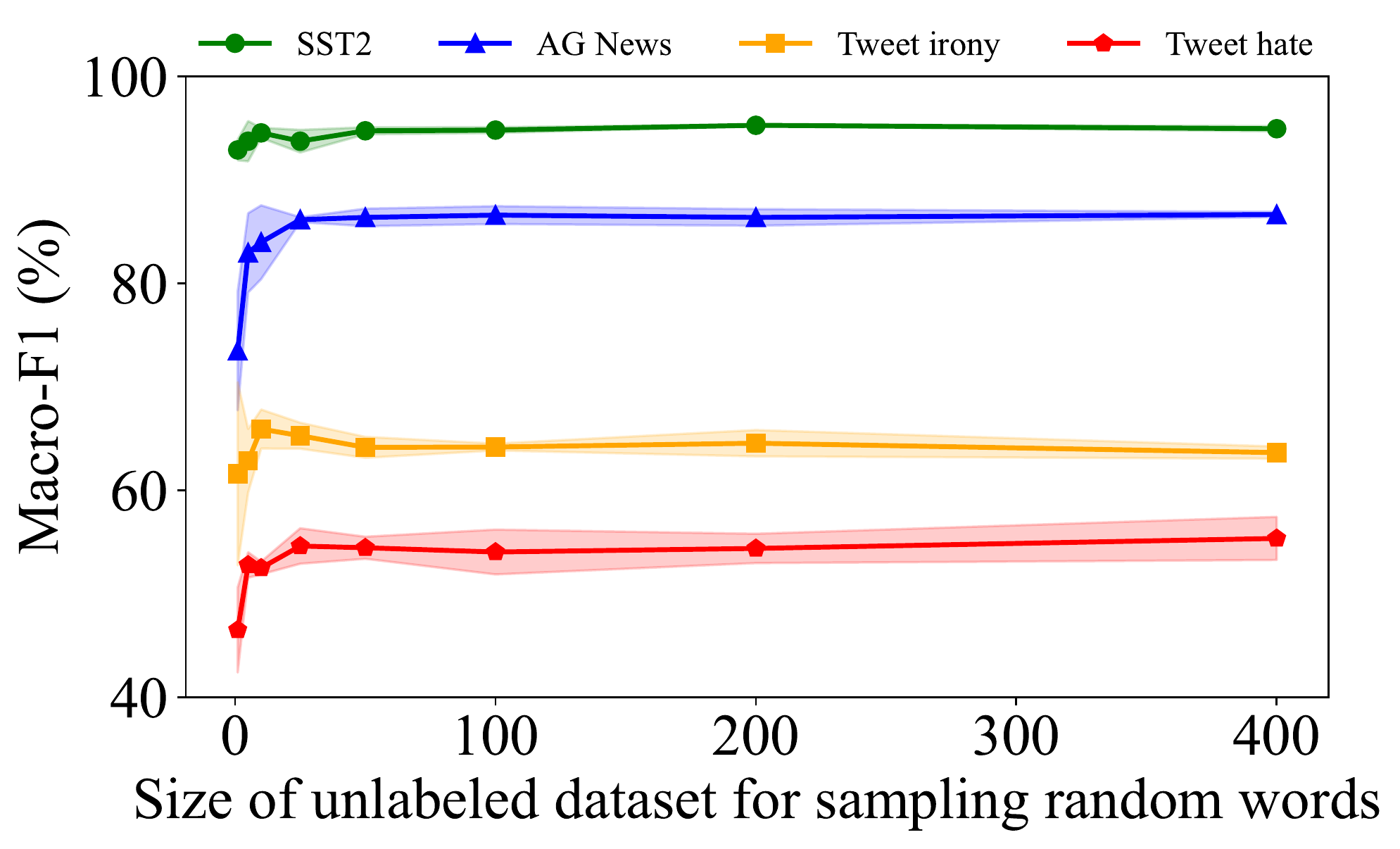}
\caption{The few-shot performance of GPT-J with DC when sampling random words from a different size of the unlabeled dataset. For each size, we sample random texts 5 times with different random seeds to show the stability of the sampling process. Using 50 unlabeled texts from the dataset already provides a stable estimate of label biases.}
\label{fig:sampling_size}
\end{figure}
\section{Zero-shot Prompting Experiment} \label{sec:zero-shot}
\paragraph{Templates for zero-shot prompting} We adapt templates from \citet{gao2020making} for our zero-shot prompting experiments.

Sentiment and detection tasks:
\begin{tcolorbox}
$\begin{aligned}
    &\langle input \rangle\ \text{It was }\langle mask \rangle\text{] } \\
\end{aligned}
$
\end{tcolorbox}

Subj:
\begin{tcolorbox}
$\begin{aligned}
    &\langle input \rangle\ \text{This is }\langle mask \rangle \\
\end{aligned}
$
\end{tcolorbox}

Topic tasks:
\begin{tcolorbox}
$\begin{aligned}
    &\langle mask \rangle \text{ : }\langle input \rangle\\
\end{aligned}
$
\end{tcolorbox}

NLI tasks:
\begin{tcolorbox}
$\begin{aligned}
    &\langle sentence_1 \rangle\ \text{ ? }\langle mask \rangle\text{ , } \langle sentence_2 \rangle \\
\end{aligned}
$
\end{tcolorbox}

\paragraph{Full Results}
We show the full zero-shot prompting results with RoBERTa-large on individual datasets in Table~\ref{tab:zeroshot_prompting}.
\newpage

\begin{table*}[t]
\centering
\begin{tabular}{lcccc}
\toprule
 \textbf{Dataset} &\textbf{\# Class} & \textbf{Balanced} &\textbf{GPT-J} &\textbf{GPT-3}\\
 \midrule
 \multicolumn{5}{l}{\emph{Sentiment and topic classification}}\\
  SST-2 \citep{socher2013recursive}             &   2   &   \cmark  &   \cmark  &   \cmark\\
  SST-5 \citep{socher2013recursive}             &   5   &   \xmark  &   \cmark  &   \cmark\\
  MR \citep{pang2005seeing}                     &   2   &   \cmark  &   \cmark  &   \\
  CR \citep{hu2004mining}                       &   2   &   \cmark  &   \cmark  &   \\
  financial\_phrasebank \citep{malo2014good}    &   3   &   \xmark  &   \cmark  &   \cmark\\
  poem\_sentiment \citep{sheng-uthus-2020-investigating} &   4   &   \xmark  &   \cmark  &   \\
  Subj  \citep{pang2004sentimental}    &   2   &   \xmark  &   \cmark  &   \\
  AG News \citep{zhang2015character}   &   4   &   \cmark  &   \cmark  &   \cmark\\
  DBpedia \citep{zhang2015character}            &   14   &   \cmark  &   \cmark  &   \\
  TREC    \citep{voorhees2000building}            &   6   &   \xmark  &   \cmark  &   \cmark\\
  \midrule
  \multicolumn{5}{l}{\emph{Natural language inference}}\\
  glue-wnli \citep{levesque2012winograd} &   2   &   \xmark  &   \cmark  &   \\
  RTE \citep{dagan2005pascal} &   2   &   \xmark  &   \cmark  &   \cmark\\
  CB  \citep{de2019commitmentbank} &   3   &   \xmark  &   \cmark  &   \cmark\\
  sick \citep{marelli2014sick}  &   3   &   \xmark  &   \cmark  &   \\
  \midrule
    \multicolumn{5}{l}{\emph{Detection}}\\
  TweetEval-hate \citep{barbieri-etal-2020-tweeteval} &   2   &   \xmark  &   \cmark  &   \cmark\\
  TweetEval-irony \citep{barbieri-etal-2020-tweeteval} &   2   &   \xmark  &   \cmark  &   \cmark\\
  TweetEval-offensive \citep{barbieri-etal-2020-tweeteval} &   2   &   \xmark  &   \cmark  &   \cmark\\
  TweetEval-stance-atheism \citep{barbieri-etal-2020-tweeteval}&   3   &   \xmark  &   \cmark  &   \\
  TweetEval-stance-feminist \citep{barbieri-etal-2020-tweeteval}  &   3   &   \xmark  &   \cmark  &   \\
  hate\_speech18 \citep{de-gibert-etal-2018-hate} &   2   &   \xmark  &   \cmark  &   \cmark\\
  ethos-binary  \citep{mollas2022ethos}  &   2   &   \xmark  &   \cmark  &   \\
  ethos-religion  \citep{mollas2022ethos} &   2   &   \xmark  &   \cmark  &   \\
  ethos-national\_origin \citep{mollas2022ethos} &   2   &   \xmark  &   \cmark  &   \\
  ethos-race  \citep{mollas2022ethos}  &   2   &   \xmark  &   \cmark  &   \cmark\\
\bottomrule
\end{tabular}
\caption{\label{tab:datasets}
Full dataset information. To control the evaluation budget, we use a subset of the 24 datasets for GPT-3 experiments following \citet{min2022rethinking}. The GPT-J and GPT-3 columns indicate whether the corresponding datasets are used in Fig.~\ref{fig:main}.
}
\end{table*}

\begin{table*}[t]
\centering
\small
\begin{tabular}{p{4cm}p{7.0cm}p{3.5cm}}
\toprule
 \textbf{Dataset} &\textbf{Template} & \textbf{Label Name}\\
 \midrule
 SST-2, SST-5, MR, CR  & \bluetext{Review: }[INPUT]\newline\bluetext{Sentiment:} [LABEL] & positive, negative\\
  \midrule
 financial phrasebank  & \bluetext{Sentence: }[INPUT]\newline\bluetext{Sentiment:} [LABEL]& positive, negative, neutral\\
  \midrule
 poem sentiment  & \bluetext{Verse text: }[INPUT]\newline\bluetext{Sentiment:} [LABEL]& positive, negative, neutral, mixed\\
 \midrule
  Subj  & \bluetext{Input: }[INPUT]\newline\bluetext{Label:} [LABEL]& subjective, objective\\
 \midrule
 AG News  & \bluetext{Article: }[INPUT]\newline\bluetext{Answer:} [LABEL] & world, sports, business, technology \& science\\
  \midrule
 DBpedia  & \bluetext{Article: }[INPUT]\newline\bluetext{Article type:} [LABEL]& company, school, artist, athlete, politics, transportation, building, nature, village, animal, plant, album, film, book\\
  \midrule
 TREC  & \bluetext{Question: }[INPUT]\newline\bluetext{Answer type:} [LABEL]& number, location, person, description, entity, abbre\\
  \midrule
 RTE, glue-wnli & [PREMISE] \bluetext{question: }[HYPOTHESIS]\newline\bluetext{True or False? answer:} [LABEL] & True, False\\
   \midrule
 CB  & [PREMISE] \bluetext{question: }[HYPOTHESIS]\newline\bluetext{true, false, or neither? answer:} [LABEL] & true, false, neither\\
   \midrule
 sick  & [PREMISE] \bluetext{question: }[HYPOTHESIS]\newline\bluetext{entailment, neutral, or contradiction? answer:} [LABEL] & entailment, neutral, contradiction\\
    \midrule
 TweetEval-hate  & \bluetext{Tweet: }[INPUT]\newline\bluetext{Label:} [LABEL]& neutral, hate\\
     \midrule
 TweetEval-irony  & \bluetext{Tweet: }[INPUT]\newline\bluetext{Label:} [LABEL]& neutral, ironic\\
     \midrule
 TweetEval-offensive  & \bluetext{Tweet: }[INPUT]\newline\bluetext{Label:} [LABEL]& neutral, offensive\\
     \midrule
 TweetEval-stance\_atheism, TweetEval-stance\_feminist   & \bluetext{Tweet: }[INPUT]\newline\bluetext{Label:} [LABEL]& none, against, favor\\
      \midrule
 hate\_speech18, ethos-race, ethos-binary, ethos-religion, ethos-national\_origin,   & \bluetext{Text: }[INPUT]\newline\bluetext{Label:} [LABEL]& neutral, hate\\
\bottomrule
\end{tabular}
\caption{\label{tab:templates}
\bluetext{Templates} of all 24 datasets used in our experiments. We mainly adapt templates used in \citet{zhao2021calibrate} and \citet{min2022rethinking}. We remove all task instructions and unify the format for similar tasks. We always use lower-case label names to avoid label names being tokenized into subwords except for ``True or False'' tasks.
}
\end{table*}
\begin{table*}[t]
\centering
\small
\begin{tabular}{p{3cm}p{11.0cm}}
\toprule
 \textbf{Dataset} &\textbf{Instruction and Template}\\
 \midrule
 TweetEval-hate  & \redtext{Classify tweets that are hateful against immigrants or women as hate and tweets that are not hateful against immigrants or women as neutral.}\newline 
 \bluetext{Tweet: }[INPUT]\newline \bluetext{Label:} [LABEL]\\
     \midrule
 TweetEval-irony  & \redtext{Classify tweets that are ironic as ironic, and tweets that are not ironic as neutral.} \newline  \bluetext{Tweet: }[INPUT]\newline\bluetext{Label:} [LABEL]\\
     \midrule
 TweetEval-offensive  & \redtext{Classify tweets that are offensive as offensive, and tweets that are not offensive as neutral.} \newline \bluetext{Tweet: }[INPUT]\newline\bluetext{Label:} [LABEL]\\
\bottomrule
\end{tabular}
\caption{\label{tab:task_instructions}
\redtext{Task instructions} used in Table~\ref{tab:instruction}.
}
\end{table*}

\begin{table*}[t]
\centering
\begin{tabular}{lllllll}
\toprule
{\bf Dataset} & \multicolumn{3}{c}{\bf GPT-J} & \multicolumn{3}{c}{\bf GPT-3} \\
& Original & CC & DC & Original & CC & DC\\
 \midrule
  \multicolumn{5}{l}{\emph{Sentiment and topic classification}}\\
SST-2 & $91.0_{6.0}$ & $90.8_{3.2}$ & ${\bf 94.0}_{1.3}$ & $96.0_{1.0}$ & ${\bf 96.4}_{0.2}$ & $96.3_{0.5}$\\
CR & $81.4_{6.9}$ & $86.5_{0.8}$ & ${\bf 87.0}_{4.0}$ & - & - & -\\
MR & $93.1_{0.7}$ & $91.3_{1.1}$ & ${\bf 93.1}_{0.5}$ & - & - & -\\
SST-5 & $28.9_{4.6}$ & ${\bf 40.8}_{5.4}$ & $40.3_{4.9}$ & $32.5_{3.3}$ & $41.1_{0.9}$ & ${\bf 42.4}_{2.9}$\\
Financial phrasebank & $46.4_{6.9}$ & $46.7_{4.2}$ & ${\bf 61.6}_{3.3}$ & $58.9_{11.6}$ & $60.6_{4.6}$ & ${\bf 69.5}_{7.1}$\\
Poem sentiment & $26.6_{6.1}$ & $25.5_{5.2}$ & ${\bf 31.4}_{3.0}$ & - & - & -\\
AG News & $68.4_{9.9}$ & $76.8_{7.2}$ & ${\bf 81.5}_{5.1}$ & $79.9_{7.0}$ & ${\bf 86.0}_{1.5}$ & $85.9_{1.1}$\\
DBpedia & $83.5_{3.0}$ & $90.6_{1.7}$ & ${\bf 92.4}_{1.2}$ & - & - & -\\
TREC & $55.3_{7.0}$ & $63.9_{1.1}$ & ${\bf 70.3}_{3.1}$ & $69.0_{13.2}$ & $76.5_{7.9}$ & ${\bf 76.9}_{7.9}$\\
Subj & $65.2_{12.2}$ & $61.9_{13.3}$ & ${\bf 70.7}_{4.3}$ & - & - & -\\
 \midrule
  \multicolumn{5}{l}{\emph{Natural language inference}}\\
RTE & $43.1_{5.8}$ & $37.8_{4.3}$ & ${\bf 50.3}_{5.6}$ & $61.8_{10.6}$ & ${\bf 65.8}_{3.8}$ & $64.5_{5.2}$\\
WNLI & $33.5_{0.0}$ & $33.9_{0.6}$ & ${\bf 38.1}_{3.1}$ & - & - & -\\
CB & $24.8_{6.1}$ & $27.8_{9.3}$ & ${\bf 42.3}_{3.6}$ & ${\bf 53.7}_{1.7}$ & $49.0_{8.4}$ & $51.1_{8.9}$\\
Sick & $25.6_{6.7}$ & $41.1_{3.7}$ & ${\bf 41.4}_{10.9}$ & - & - & -\\
 \midrule
    \multicolumn{5}{l}{\emph{Detection}}\\
TweetEval-hate & $32.8_{1.2}$ & $36.4_{3.4}$ & ${\bf 61.2}_{2.3}$ & $36.8_{4.7}$ & $49.5_{7.6}$ & ${\bf 59.0}_{4.1}$\\
TweetEval-irony & $60.0_{10.4}$ & $50.3_{7.7}$ & ${\bf 62.4}_{5.3}$ & $42.7_{14.1}$ & $37.5_{10.6}$ & ${\bf 62.7}_{7.6}$\\
TweetEval-offensive & $59.4_{11.4}$ & $51.0_{6.3}$ & ${\bf 68.3}_{2.9}$ & $59.3_{6.1}$ & $60.5_{3.6}$ & ${\bf 64.8}_{1.9}$\\
TweetEval-stance-atheism & $23.2_{5.1}$ & $23.0_{6.0}$ & ${\bf 27.4}_{3.2}$ & - & - & -\\
TweetEval-stance-feminist & $40.4_{10.0}$ & $34.6_{2.6}$ & ${\bf 41.3}_{1.9}$ & - & - & -\\
Hate speech18 & $51.5_{4.7}$ & $41.6_{8.6}$ & ${\bf 57.3}_{2.5}$ & $49.9_{11.5}$ & $47.0_{6.2}$ & ${\bf 52.0}_{3.9}$\\
Ethos binary & $48.4_{16.1}$ & $60.1_{8.2}$ & ${\bf 70.2}_{2.5}$ & - & - & -\\
Ethos religion & $30.7_{14.3}$ & $28.0_{13.8}$ & ${\bf 43.8}_{6.7}$ & - & - & -\\
Ethos nation & $23.1_{8.7}$ & $18.2_{2.1}$ & ${\bf 40.7}_{7.8}$ & - & - & -\\
Ethos race & $36.4_{11.8}$ & $44.8_{17.4}$ & ${\bf 51.4}_{6.4}$ & $40.0_{12.2}$ & $33.2_{7.6}$ & ${\bf 47.0}_{6.9}$\\
\bottomrule
\end{tabular}
\caption{\label{tab:full}
Full 8-shot results. We report the average Macro-F1 scores over 5 random seeds with the standard deviations. To control the evaluation budget, we use a subset of the 24 datasets for GPT-3 experiments following \citet{min2022rethinking}.
}
\end{table*}

\begin{table*}[t]
\centering
\small
\begin{tabular}{lllllllllllll}
\toprule
{\bf Setting} & {\bf SST-2} & {\bf CR} & {\bf MR} & {\bf SST-5}  & {\bf FP} & {\bf PS} & {\bf AG} & {\bf DB} & {\bf TREC} & {\bf Subj} & {\bf RTE} & {\bf WNLI}\\
 \midrule
Ori.& 82.5& 82.2& 78.5& 28.3& 32.5& 24.1& 52.0& 65.5& 7.0& 37.9& 39.2& 32.9\\
CC& 79.4& 81.0& 76.0& 19.3& 40.0& 21.7& 59.1& 73.1& 5.0& {\bf 51.6}& {\bf 46.0}& 33.6\\
DC& {\bf 87.2}& {\bf 82.2}& {\bf 82.5}& {\bf 35.1}& {\bf 43.3}& {\bf 26.2}& {\bf 62.7}& {\bf 74.2}& {\bf 22.6}& 44.3& 44.6& {\bf 33.8}\\
 \midrule
 {\bf Setting} & {\bf Tw-H} & {\bf Tw-I} & {\bf Tw-O} & {\bf Tw-A}  & {\bf Tw-F} & {\bf HS18} & {\bf Eth-B} & {\bf Eth-Re} & {\bf Eth-N} & {\bf Eth-Ra} & {\bf CB} & {\bf Sick}\\
 \midrule
Ori.& 32.9& 30.5& 28.4& 23.7& 22.4& 37.7& 45.0& 21.0& 17.8& 17.3& 21.3& 38.2\\
CC& 30.5& 30.8& 32.8& 19.5& 28.0& 26.5& 37.0& 17.6& 15.5& 15.8& {\bf 34.9}& 39.7\\
DC& {\bf 59.0}& {\bf 49.7}& {\bf 56.3}& {\bf 28.5}& {\bf 29.4}& {\bf 44.3}& {\bf 58.7}& {\bf 42.7}& {\bf 36.9}& {\bf 44.1}& 27.5& {\bf 54.2}\\

\bottomrule
\end{tabular}
\caption{\label{tab:zeroshot_prompting}
Zero-shot prompting results with RoBERTa-large (Macro-F1).
}
\end{table*}

\end{document}